%% file: neurips_2025.tex
\documentclass{article}

\usepackage[preprint]{neurips_2025}

\usepackage{amssymb}
\usepackage{adjustbox}
\usepackage{tabularx}
\usepackage{nicefrac}       % compact symbols for 1/2, etc.
\usepackage{microtype}      % microtypography
\usepackage[dvipsnames]{xcolor}       % colors
\usepackage{graphicx}
\usepackage{subfigure}
\usepackage{multirow}
\usepackage{wrapfig}
\usepackage{enumitem}
\usepackage{algorithm}
\usepackage{algorithmic} 
\usepackage[utf8]{inputenc} % allow utf-8 input
\usepackage[T1]{fontenc}    % use 8-bit T1 fonts
\usepackage{hyperref}       % hyperlinks
\usepackage{url}            % simple URL typesetting
\usepackage{booktabs}       % professional-quality tables
\usepackage{amsfonts}       % blackboard math symbols
\usepackage{nicefrac}       % compact symbols for 1/2, etc.
\usepackage{microtype}      % microtypography
\usepackage{xcolor}         % colors
 \usepackage{amsmath}
\title{TarDiff: Target-Oriented Diffusion Guidance \\ for Synthetic Electronic Health Record \\ Time Series Generation}

\author{%
  Bowen Deng$\textsuperscript{1}$\thanks{The work was conducted during the internship of Bowen Deng and Hao Li at Microsoft Research.},\hspace{0.25em}
  Chang Xu$\textsuperscript{2}$\thanks{Corresponding author.}, \hspace{0.25em}
  Hao Li$\textsuperscript{3}$,\hspace{0.25em}
  Yuhao Huang$\textsuperscript{4}$,\hspace{0.25em}
  Min Hou$\textsuperscript{5}$,
  Jiang Bian$\textsuperscript{2}$ \\
  $\textsuperscript{1}$ Peking University \hspace{0.25em}
  $\textsuperscript{2}$ Microsoft Research Asia \hspace{0.25em}
  $\textsuperscript{3}$ University of Manchester\\
  $\textsuperscript{4}$ Nanjing University \hspace{0.25em}
  $\textsuperscript{5}$ Hefei University of Technology \\
  devin@stu.pku.edu.cn \hspace{0.25em}
  \{chanx, jiang.bian\}@microsoft.com \hspace{0.25em} \\
  hao.li-2@manchester.ac.uk \hspace{0.25em}
  huangyh@smail.nju.edu.cn  \hspace{0.25em} \\
  hmhoumin@gmail.com \hspace{0.25em}
}

\begin{document}

\maketitle

\input{section/abstract}
\input{section/intro}
\input{section/preliminary}
\input{section/method}
\input{section/Experiment}
\input{section/related_work}

\input{section/conclusion}

% \textcolor{blue} {Why was precisely what we done necessary to address the problem outlined in the introduction? Every methodological step taken needs to be justified with this in mind.}

% \textcolor{blue} {Concept and implementation need to be separated here. In the paper we talk about concepts/ideas and explain why these, whereas implementation details might either go into the appendix or be left out completely.}

\bibliographystyle{plainnat}
\bibliography{NIPS}

%%%%%%%%%%%%%%%%%%%%%%%%%%%%%%%%%%%%%%%%%%%%%%%%%%%%%%%%%%%%

\appendix
\input{section/appendix}

\end{document}

%% file: section/abstract.tex
\begin{abstract}

Synthetic Electronic Health Record (EHR) time-series generation is crucial for advancing clinical machine learning models, as it helps address data scarcity by providing more training data.
However, most existing approaches focus primarily on replicating statistical distributions and temporal dependencies of real-world data. 
We argue that fidelity to observed data alone does not guarantee better model performance, as common patterns may dominate, limiting the representation of rare but important conditions. 
This highlights the need for generate synthetic samples to improve performance of specific clinical models to fulfill their target outcomes.
To address this, we propose \textit{TarDiff}, a novel target-oriented diffusion framework that integrates task-specific influence guidance into the synthetic data generation process. 
Unlike conventional approaches that mimic training data distributions, TarDiff optimizes synthetic samples by quantifying their expected contribution to improving downstream model performance through influence functions.
Specifically, we measure the reduction in task-specific loss induced by synthetic samples and embed this influence gradient into the reverse diffusion process, thereby steering the generation towards utility-optimized data.
Evaluated on \textbf{six} publicly available EHR datasets, TarDiff achieves state-of-the-art performance, outperforming existing methods by up to 20.4\% in AUPRC and 18.4\% in AUROC.
Our results demonstrate that TarDiff not only preserves temporal fidelity but also enhances downstream model performance, offering a robust solution to data scarcity and class imbalance in healthcare analytics. 

\end{abstract}

%% file: section/intro.tex
\section{Introduction}

Healthcare is a cornerstone of societal well-being, especially as the world faces an aging population and the rising burden of chronic diseases, placing increasing pressure on healthcare systems worldwide \citep{liang2024user}. 
Traditionally, medical diagnoses have relied on human expertise, but with advancements in machine learning, Electronic Health Records (EHRs)—which digitally store a patient’s medical history, including demographic attributes \citep{DBLP:journals/tim/MaweuSDP21,DBLP:conf/bionlp/LiWSBNKZB0N23}, vital signs \citep{tseng2022fast}, and lab measurements—have become invaluable for clinical research \citep{DBLP:journals/fdata/KaushikCSDNPD20, DBLP:conf/clef/SchlegelLW0NKBZ23}. 
EHR time series data, such as Electroencephalography (EEG) for neurological analysis and Electrocardiography (ECG) for heart condition diagnosis, provide critical insights for medical decision-making.  
Leveraging these rich data sources, machine learning models trained in a data-driven manner are then applied to various downstream tasks, including disease diagnosis, prognosis prediction, and treatment planning~\citep{shickel2017deep,goldstein2016opportunities,DBLP:conf/acl/LiZQLLWLYMZZLZM24,DBLP:journals/corr/abs-2408-12249}.

However, obtaining and utilizing EHR data remains a significant challenge due to medical-related factors, such as strict privacy regulations, data incompleteness resulting from sensor failures, and difficulties in accurate labeling.
As a result, synthesizing EHR data has gained increasing attention. 
Existing work has explored a wide range of methods including rule-based approaches and generative models. 
Rule-based techniques—such as time warping, jittering, and interpolation\citep{medformer,traditionalaug}—are favored for their simplicity and efficiency, yet they often fail to capture the intricate temporal dependencies and pathological patterns present in clinical data. 
GANs \citep{DBLP:journals/corr/abs-2301-05465} and VAEs~\citep{kingma2013auto} have been employed to generate high-fidelity signals, with notable examples including TimeGANs~\citep{timegan}, TimeVAE~\citep{timevae}, and TimeVQ-VAE~\citep{timevqvae}, which have demonstrated improved performance in various biohealthcare applications. %\bowen{In essence, VAEs aim to learn a latent space that captures the intrinsic distribution of the training data, thus enabling the generation of realistic synthetic samples.}
More recently, diffusion models~\citep{ho2020denoising,song2020score,DBLP:conf/iclr/FanW0HL024,li2025bridge} have emerged as a promising alternative; methods such as TimeDiff~\citep{private}, DiffusionTS~\citep{diffusionts} and BioDiffusion~\citep{DBLP:journals/corr/abs-2401-10282} similarly seek to approximate the underlying data distribution through iterative refinement, emphasizing the generation of realistic time series data.

While these efforts have significantly advanced the field, they primarily focus on generating synthetic data by mimicking the empirical distribution of training samples.
We argue that this approach inherently overlooks the effectiveness of the generated data for downstream medical tasks. 
For instance, in rare disease diagnosis, where positive samples are scarce in real-world datasets, generating synthetic data purely based on observed distributions may exacerbate biases in the downstream model~\citep{syntbias}, making it more inclined to diagnose common conditions while failing to recognize rare diseases effectively. 
Consequently, such data generation strategies do not necessarily enhance model performance in rare disease diagnosis and may even worsen the imbalance in predictive accuracy~\citep{imbalancegen}. 
Therefore, we propose that EHR data generation should not merely replicate statistical patterns of training data but should instead be guided by its utility in training more effective models to fulfill the targets of downstream tasks. 
This calls for an adaptive generation strategy that actively optimizes synthetic data to enhance model learning for specific medical applications.

In this work, we investigate how to develop a model for generating EHR data that is specifically tailored to enhance downstream model performance for target tasks. 
Recent advancements in diffusion models have shown promising results in time series generation~\citep{private,diffusionts,DBLP:journals/corr/abs-2401-10282}, with conditional diffusion models~\citep{timedp,dhariwal2021diffusion,ho2022classifier} offering significant insights for this work. These models have the ability to guide the generation process towards a predefined goal, which provides a compelling direction for EHR generation.
Therefore, an intuitive idea is to guide the diffusion process towards generating data that benefits downstream tasks. 
However, one of the key challenges is how to represent and quantify the impact of the generated data on the performance of downstream models for specific tasks. 
To address this, we draw inspiration from influence functions\citep{DBLP:conf/icml/KohL17,anand2023influence, cook1977detection}, a robust statistics technique that measures the impact of a single data point on an estimator, revealing how observations affect model parameters and predictions. In machine learning, influence functions help understand model behavior, debug predictions, and identify influential training points by tracing a model’s predictions back to its training data \citep{DBLP:journals/corr/abs-2412-08480}.

Building upon this foundation, we propose a diffusion framework that integrates the influence of synthetic samples as a form of guidance, with the goal of generating samples that yield the most positive impact on specific clinical tasks. In our approach, the influence of a generated sample is defined as the reduction in the task-specific loss on a guidance set drawn from the same distribution as the downstream task when the sample is incorporated into the training data. By estimating the influence of intermediate samples during the diffusion process, we leverage the gradient information associated with the influence to steer the generation process toward producing data that are optimally beneficial for downstream tasks.
By incorporating this influence gradient into the reverse diffusion process, our method actively guides the generation toward producing samples that are more likely to improve performance of healthcare prediction tasks. 
%For example, synthetic ECG signals intended for arrhythmia detection must retain features critical for accurate diagnosis—a requirement that is directly enforced by maximizing the influence measure. 
Ultimately, the influence mechanism bridges the gap between data authenticity and clinical utility, ensuring that the generated time series are not only realistic but also tailored to enhance specific healthcare tasks.

To summarize, this paper makes the following contributions: First, we introduce a novel influence mechanism that quantifies the clinical utility of synthetic samples by measuring the expected reduction in task-specific loss. Second, we propose TarDiff, an influence guided diffusion model that integrates this mechanism into the reverse diffusion process, effectively steering generation toward samples that enhance downstream model performance. 
Third, we validate our approach on multiple clinical tasks, demonstrating that the synthetic medical time series produced by our method are not only realistic but also yield significant improvements in clinical outcomes.

%% file: section/preliminary.tex
\section{Preliminary}
\subsection{Diffusion Models}\label{sec:prelim-guidance}

\paragraph{Denoising Diffusion Probabilistic Models}

The core idea behind DDPM\citep{ho2020denoising} is the modeling of a forward diffusion process, which progressively adds noise to the data, and a reverse diffusion process, which denoises the data to ultimately generate realistic samples from noise. Let \( x_0 \) represent a data sample drawn from the real data distribution \( p_{\text{data}}(x_0) \). The forward diffusion process introduces Gaussian noise to the data over \( T \) discrete time steps, transforming \( x_0 \) into pure noise \( x_T \), which is assumed to follow a standard normal distribution, i.e., \( x_T \sim \mathcal{N}(0, I) \). At each time step \( t \), the data \( x_t \) is modeled as a noisy version of \( x_{t-1} \) via the conditional Gaussian distribution:
\begin{equation}
\begin{split}
q(x_t | x_{t-1}) &= \mathcal{N}(x_t; \sqrt{1 - \beta_t} x_{t-1}, \beta_t I), \\
q(x_t | x_0) &= \mathcal{N}(x_t; \sqrt{\bar{\alpha}_t} x_0, (1 - \bar{\alpha}_t) I).
\end{split}
\end{equation}
 \( \beta_t \) denotes the noise schedule controlling the amount of noise added at each step, and \( \bar{\alpha}_t = \prod_{s=1}^t (1 - \beta_s) \) represents the cumulative noise factor. The forward diffusion process is Markovian, and the joint distribution of the noisy data can be expressed as \( q(x_{1:T} | x_0) = \prod_{t=1}^{T} q(x_t | x_{t-1}) \), where \( x_{1:T} = (x_1, x_2, ..., x_T) \) represents the sequence of noisy variables from \( x_0 \) to \( x_T \).

The reverse diffusion process seeks to recover the original data \( x_0 \) from the noise \( x_T \) by learning a generative model. Specifically, at each time step, the model predicts the mean of the reverse distribution \( p_\theta(x_{t-1} | x_t) \), which is also assumed to be Gaussian:
\begin{equation}
\begin{split}
p_\theta(x_{t-1} | x_t) &= \mathcal{N}(x_{t-1}; \mu_\theta(x_t, t), \Sigma_\theta(t)),
\end{split}
\end{equation}
where \( \mu_\theta(x_t, t) \) and \( \Sigma_\theta(t) \) represent the model-predicted mean and covariance, respectively, parameterized by \( \theta \).

The model is trained by minimizing the following loss function, leveraging the Markov property of the diffusion process:
\begin{equation}
L(\theta) = \mathbb{E}_q \left[ \| \hat{\epsilon}_\theta(x_t, t) - \epsilon_t \|^2 \right],
\end{equation}
 \( \hat{\epsilon}_\theta(x_t, t) \) is the model’s prediction of the noise added at time step \( t \), and \( \epsilon_t \) is the actual noise introduced during the forward diffusion process. This loss function encourages the model to accurately predict the added noise, enabling it to reverse the diffusion process and recover the original data \( x_0 \).
\paragraph{Conditional Diffusion}
In conditional diffusion\citep{ho2022classifier}, the reverse process is explicitly guided by an additional condition $y$(i.e.,class label). Specifically, the probability is modeled as:
\begin{equation}\label{conditiondf}
    p_\theta(x_{t-1} | x_t) = \mathcal{N}(x_{t-1}; \mu_\theta(x_t, y, t), \Sigma_\theta(t)),
\end{equation}
where the mean function $\mu_\theta(x_t, y, t)$ incorporates not only the current state $x_t$ and the time step $t$ but also the condition $y$, and $\Sigma_\theta(t)$ denotes the covariance at time $t$. This formulation enables the model to generate samples that adhere to both the learned data distribution and the desired attributes specified by $y$, thereby achieving controlled synthesis.

In the following, we adapt the conditional diffusion framework to time series generation. To make this concrete, we first define the general time series generation problem.

\paragraph{Classifier-Guided Diffusion.}
This approach\citep{dhariwal2021diffusion} augments reverse diffusion with signals from an classifier that estimates the conditional likelihood \(p(y\mid x_t)\) at every timestep \(t\).
The gradient
\(\nabla_{x_t}\log p(y\mid x_t)\)
indicates the direction in sample space that most increases the probability of label \(y\);
adding this vector therefore nudges the denoising trajectory toward regions consistent with the desired condition.

Formally, the mean of the Gaussian transition in Eq.~\eqref{conditiondf} is replaced by
\begin{equation}
\tilde{\mu}_\theta(x_t,y,t)
\;=\;
\mu_\theta(x_t,y,t)
\;+\;
\alpha\,\nabla_{x_t}\log p(y\mid x_t),
\end{equation}
where the scalar \(\alpha\) controls guidance strength.
The resulting reverse kernel becomes
\begin{equation}
p_\theta(x_{t-1}\mid x_t)
=\mathcal{N}\!\bigl(x_{t-1};\,
\tilde{\mu}_\theta(x_t,y,t),\,
\Sigma_\theta(t)\bigr).
\end{equation}

By favouring states that yield larger \(\log p(y\mid x_t)\), classifier guidance systematically steers the generative process toward samples that satisfy the target label, compensating for mismatches in the diffusion model’s original conditional distribution.

\subsection{Task Formulation}

Let $\mathbf{X} = \{\mathbf{x}_1, \mathbf{x}_2, \dots, \mathbf{x}_T\}$ be a continuous time series of length $T$, where each observation $\mathbf{x}_t \in \mathbb{R}^d$. 
Suppose we have a dataset $\mathcal{D}_0 = \{\mathbf{X}^{(i)}\}_{i=1}^{N}$, composed of $N$ independent sequences drawn from an unknown underlying distribution 
\begin{equation}
    p_{\text{data}}(\mathbf{X}) \;=\; p_{\text{data}}\bigl(\mathbf{x}_1, \mathbf{x}_2, \dots, \mathbf{x}_T\bigr).
\end{equation}
Our goal is to learn a generative model $p_{G}(\mathbf{X})$ that approximates $p_{\text{data}}(\mathbf{X})$, enabling us to sample new sequences
\begin{equation}
    \hat{\mathbf{X}} = \{\hat{\mathbf{x}}_1, \hat{\mathbf{x}}_2, \ldots, \hat{\mathbf{x}}_T\}
    \quad \text{with} \quad
    \hat{\mathbf{X}} \sim p_{G}(\mathbf{X}).
\end{equation}

A common strategy is to minimize some divergence measure between $p_{\text{data}}$ and $p_G$,
\begin{equation}
    \min_{p_G} \, D\bigl(p_{\text{data}}(\mathbf{X}) \,\big\|\, p_G(\mathbf{X})\bigr),
\end{equation}
where $D(\cdot\|\cdot)$ could be the KL divergence. In practice, $p_G$ must capture both local dependencies (e.g., between adjacent time steps) and global trends. Let $S(\mathbf{X})$ denote relevant statistics (e.g., autocorrelation or cross-correlation). Then a suitable generative model should satisfy
\begin{equation}
    \mathbb{E}_{\hat{\mathbf{X}} \sim p_G}\bigl[S(\hat{\mathbf{X}})\bigr]
    \;\approx\;
    \mathbb{E}_{\mathbf{X} \sim p_{\text{data}}}\bigl[S(\mathbf{X})\bigr],
\end{equation}
so that synthetic sequences reflect the essential temporal structures of real data.

For conditional diffusion models, the generative process can be guided by incorporating conditional information $y$, such as class labels, as formulated in Equation~\eqref{conditiondf}. This conditioning mechanism allows the model to generate time series that align with specific contexts or constraints while preserving both local dependencies and global trends.

%% file: section/method.tex
\section{Methodology}
\begin{figure*}[t]
    \centering
      \includegraphics[width=\textwidth]{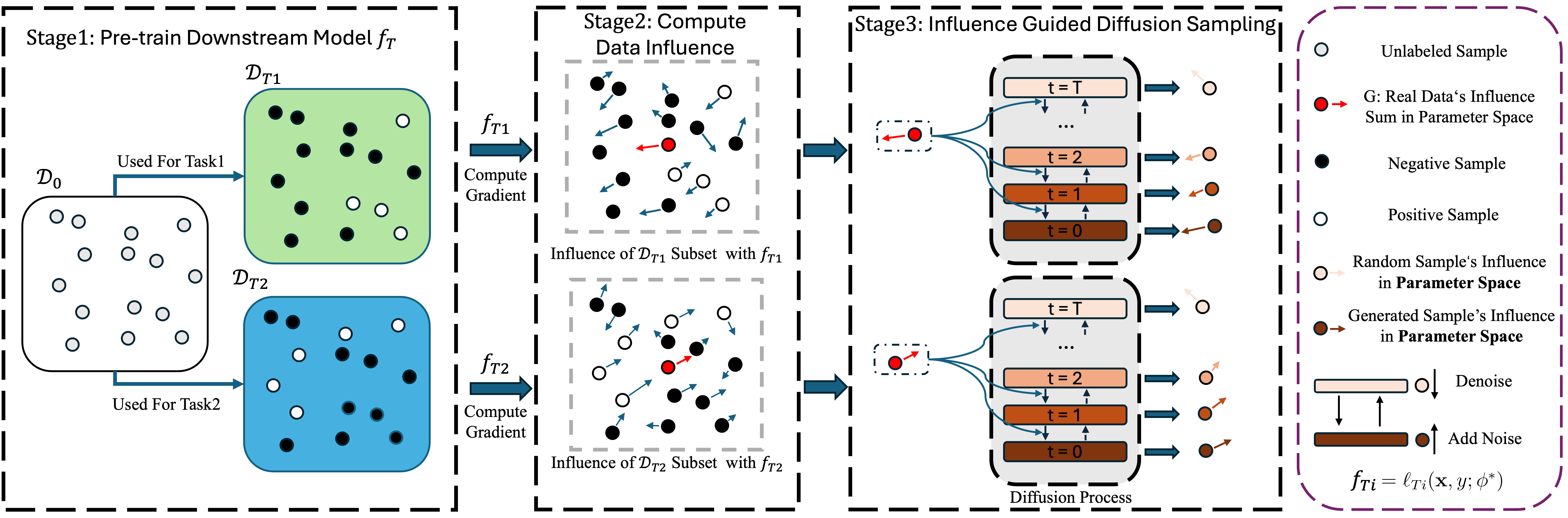}
        \caption{
        \textbf{Overview of the Influence Guidance Diffusion framework.}
        In \textbf{Stage 1}, we construct task-specific datasets from the original dataset $\mathcal{D}_{train}$ and train downstream models $f_{T_i}$
        In \textbf{Stage 2}, we compute each sample's gradient-based influence for total influence $\mathbf{G}$ based on $\mathcal{D}_{T_i}$ and $f_{T_i}$.
        In \textbf{Stage 3}, we leverage influence signals guide the reverse diffusion process with computing $\Delta \mathcal{L}_{T}(\hat{z}) =   \nabla_{\phi}  \ell_T(
        \mathbf{x}_t,y_t; \phi)\cdot{G}$. All symbols are detailed in the legend on the right.
        }
      \label{fig:framework}
\end{figure*}

In this section, we describe our TarDiff framework in detail. As can be seen from Figure~\ref{fig:framework}, our method builds upon a conditional diffusion model to generate useful synthetic data through explicitly incorporating task-specific influence signals into the reverse diffusion process.

\subsection{Influence Formulation}\label{pf}
We consider a training dataset $\mathcal{D}_{\text{train}}=\{(\mathbf{x}_i,y_i)\}_{i=1}^{n}$ and a collection of downstream tasks $\mathcal{T}=\{T_1,T_2,\dots\}$.  
For any task $T\in\mathcal{T}$, the task‑specific parameters are obtained by minimising the empirical loss on $\mathcal{D}_{\text{train}}$:
\begin{equation}
\phi^{*}_{T}
=
\arg\min_{\phi}\;
\sum_{(\mathbf{x}_i,y_i)\in\mathcal{D}_{\text{train}}}
\ell_{T}\!\bigl(\mathbf{x}_i,y_i;\phi\bigr).
\end{equation}

To mitigate limitations arising from insufficient or noisy training data, we generate a synthetic sample $\hat{z}=(\mathbf{x},y)$ and augment the original data, yielding  
$\mathcal{D}_{0}\cup\{\hat{z}\}$, where $\mathcal{D}_{0}\!:=\!\mathcal{D}_{\text{train}}$.  
Retraining on the augmented set gives
\begin{equation}
\phi^{\hat{z}}_{T}
=
\arg\min_{\phi}\;
\sum_{(\mathbf{x}_i,y_i)\in\mathcal{D}_{0}\cup\{\hat{z}\}}
\ell_{T}\!\bigl(\mathbf{x}_i,y_i;\phi\bigr).
\end{equation}

For a samples $(\mathbf{x}',y')$, we define
\begin{equation}\label{eq:def-H}
H_{\phi}\!\bigl(\mathbf{x},y,\mathbf{x}',y'\bigr)
\;=\;
\ell_{T}\!\bigl(\mathbf{x}',y';\phi^{\hat{z}}\bigr)
-
\ell_{T}\!\bigl(\mathbf{x}',y';\phi^{*}_{T}\bigr),
\end{equation}
the change in downstream loss on $(\mathbf{x}',y')$ caused by adding $(\mathbf{x},y)$ to the training set.\footnote{%
Throughout, $\phi^{\hat{z}}$ depends on $(\mathbf{x},y)$ as in Eq.~(2), and $\phi^{*}_{T}$ is the original optimum in Eq.~(1).}

Let $\mathcal{P}$ denote the underlying data‑generating distribution that produces unseen i.i.d.\ samples at evaluation time. 
We define the \emph{influence} of $\hat{z}$ on task $T$ as the expected reduction in loss over this distribution:

\begin{equation}\label{eq:ifdefine}
\Delta\mathcal{L}_{T}(\hat{z})
\;\triangleq\;
-
\mathbb{E}_{(\mathbf{x}',y')\sim\mathcal{P}}
\bigl[
H_{\phi}\!\bigl(\mathbf{x},y,\mathbf{x}',y'\bigr)
\bigr]
=
-
\mathbb{E}_{(\mathbf{x}',y')\sim\mathcal{P}}
\Bigl[
\ell_{T}\!\bigl(\mathbf{x}',y';\phi^{\hat{z}}\bigr)
-
\ell_{T}\!\bigl(\mathbf{x}',y';\phi^{*}_{T}\bigr)
\Bigr].
\end{equation}

Our goal is to synthesise the sample that maximises this influence:
\begin{equation}
\hat{z}^{*}
=
\arg\max_{\hat{z}}\;
\Delta\mathcal{L}_{T}(\hat{z}),
\end{equation}
thereby ensuring that the generated data yields the greatest expected performance gain on unseen, i.i.d.\ instances of the target task.

\subsection{Influence Guidance Diffusion}\label{IGD}

While our pre-trained conditional diffusion model with parameters $\theta^*$ is adept at generating realistic medical time series dataset $\mathcal{D}_0$, its generation is driven solely by the learned data distribution conditioned on $y$ (e.g., the label), with the reverse diffusion process described by Equation \eqref{conditiondf}. In this standard setting, to generate a time series sample with a specific label, one simply samples from Gaussian noise and iteratively denoises according to \eqref{conditiondf}---each step conditioned only on the time step $t$ and the label $y$---to yield data consistent with the designated label.

Section~\ref{sec:prelim-guidance} reviewed how an auxiliary classifier steers the reverse kernel via the gradient $\nabla_{x_t}\log p(y\mid x_t)$, yielding the modified mean
$\tilde{\mu}_{\theta}(x_t,y,t)=\mu_{\theta}(x_t,y,t)+\alpha\nabla_{x_t}\log p(y\mid x_t)$.

Inspired by classifier guidance, we incorporate an additional control signal that measures the impact of a synthetic sample on downstream performance.
Let $\mathcal{D}_{guide}=\{(\mathbf{x}'_j,y'_j)\}_{j=1}^{N_g}$ be a \textbf{guidance set} drawn i.i.d.\ from the same data-generating distribution $\mathcal{P}$ introduced in Section~\ref{pf}.  

Recalling the point-wise loss change $H_{\phi}(\cdot)$ defined in Eq.~\eqref{eq:def-H},  
the influence of a sample $\hat{z}=(\mathbf{x},y)$ on task~$T$ is estimated by

\begin{equation}\label{eq:origin}
\Delta\mathcal{L}_{T}(\hat{z})
\;=\;
\sum_{(\mathbf{x}',y')\in\mathcal{D}_{guide}}
H_{\phi}\bigl(\mathbf{x},y,\mathbf{x}',y'\bigr).
\end{equation}

At each diffusion step $t$ we treat
\(
\hat{z}_t=(x_t,y)
\)
and replace the classifier-guidance term
$\nabla_{x_t}\log p(y\mid x_t)$
with the gradient of the influence estimate,
$\nabla_{x_t}\Delta\mathcal{L}_{T}(\hat{z}_t)$.
Note that the constant factor in Eq.~\eqref{eq:origin} is omitted, as it does not affect the direction of the gradient used for guidance.

Consequently, the reverse update becomes
\begin{equation}\label{eq:modified-guidance}
\tilde{\mu}_{\theta}(x_t,y,t)
=
\mu_{\theta}(x_t,y,t)
\;+\;
\alpha\,
\nabla_{x_t}\Delta\mathcal{L}_{T}(\hat{z}_t),
\end{equation}
where $\alpha$ controls guidance strength.

By steering the denoising trajectory toward samples that exhibit higher $\Delta\mathcal{L}_{T}(\hat{z})$, we ensure that the generated medical time series data not only conform to the specified condition label but also actively enhance downstream task performance across the guidance set. This approach mitigates spurious correlations that can arise from purely label-conditioned diffusion, as the model explicitly seeks synthetic samples that yield beneficial effects on a broad set of guidance set examples.

\subsection{Estimates of Influence} 
In \ref{pf} and \ref{IGD}, we defined the concept of influence for synthesized samples and discussed how to generate samples with the highest possible influence during the generation process. Building on this foundation, the generation of task-specific medical time series requires evaluating the influence of a generated sample on downstream tasks. The primary challenge is to efficiently quantify how synthetic samples impact the model’s parameter updates, which are designed to minimize the task-specific loss function $\ell_{T}$.

A straightforward but computationally expensive approach would involve iteratively adding each candidate synthetic sample $\hat{z}$ to the training set, retraining the model from scratch, and measuring the resultant performance change. Since this process requires $\mathcal{O}(n)$ retraining steps, it leads to prohibitive computational costs, particularly for large-scale datasets and complex models. Thus, optimizing for both efficiency and accuracy in influence estimation is crucial to enable practical applications of this method.

To circumvent the need for exhaustive retraining, a framework was proposed in \citep{graddot,anand2023influence} that approximates the parameter shift \(\delta \phi\) induced by a synthetic sample \(\hat{z}\) via gradient-based analysis, thereby significantly reducing computational overhead while maintaining accuracy in influence estimation.

When estimating the impact of a single sample on model parameters, we build upon the classic conclusion presented in \citep{graddot}. As derived in \citep{graddot}, to change the value of the model's prediction on a sample $\mathbf{x}$, denoted as $f_{\phi}(\mathbf{x})$, by a small quantity $\varepsilon$, The parameters $\phi$ can be updated according to $\delta \phi = \frac{\varepsilon \nabla_{\phi} f_{\phi}(\mathbf{x})}{\|\nabla_{\phi} f_{\phi}(\mathbf{x})\|^2}.$
After this parameter update, the new prediction at $\mathbf{x}$, $f_{\phi + \delta \phi}(\mathbf{x})$is given by:
\begin{equation}
\begin{split}
f_{\phi + \delta \phi}(\mathbf{x}) &= f_{\phi}(\mathbf{x}) + \nabla_{\phi} f_{\phi}(\mathbf{x}) \cdot \delta \phi + \mathcal{O}(\|\delta \phi\|^2) \\
&= f_{\phi}(\mathbf{x}) + \varepsilon + \mathcal{O}(\varepsilon^2)
\end{split}
\end{equation}
where $\mathcal{O}(\|\delta \phi\|^2)$ denotes higher-order terms that are negligible for sufficiently small updates. This foundational perspective allows us to analyze the impact of individual samples on model behavior without necessitating full retraining.
Furthermore, we can assess how such a parameter change affects the model's prediction on another sample $\mathbf{x}'$:
\begin{equation}
\begin{split}
    f_{\phi + \delta \phi}(\mathbf{x}') = f_{\phi}(\mathbf{x}') + \nabla_{\phi} f_{\phi}(\mathbf{x}') \cdot \delta \phi + \mathcal{O}(\|\delta \phi\|^2) \\ = 
    f_{\phi}(\mathbf{x}') + \varepsilon \frac{\nabla_{\phi} f_{\phi}(\mathbf{x}') \cdot \nabla_{\phi} f_{\phi}(\mathbf{x})}{\|\nabla_{\phi} f_{\phi}(\mathbf{x})\|^2} + \mathcal{O}(\|\delta \phi\|^2).
\end{split}
\end{equation}
The term \(\varepsilon \frac{\nabla_{\phi} f_{\phi}(\mathbf{x}') \cdot \nabla_{\phi} f_{\phi}(\mathbf{x})}{\|\nabla_{\phi} f_{\phi}(\mathbf{x})\|^2}\) quantifies the influence of the parameter update caused by $\mathbf{x}$ on the model's prediction at $\mathbf{x}'$. Specifically, the numerator $\nabla_{\phi} f_{\phi}(\mathbf{x}') \cdot \nabla_{\phi} f_{\phi}(\mathbf{x})$ represents the alignment between the gradients of $f_{\phi}$ at $\mathbf{x}'$ and $\mathbf{x}$. A higher alignment indicates that changes in $\phi$ due to $\mathbf{x}$ will have a more pronounced effect on the prediction at $\mathbf{x}'$.
\paragraph{Influence on Performance.}To extend this analysis to our setting, we replace the model's prediction $f_{\phi}(\mathbf{x})$ with a target loss function $\ell_{\text{T}}(\mathbf{x}, y; \phi)$, transitioning from analyzing the influence on predictions to quantifying the influence on the overall optimization objective. Specifically, the influence of $\mathbf{x}$ on the loss at another sample $\mathbf{x}'$ can be expressed as:
\begin{equation}
\begin{split}
    \ell_{\text{T}}(\mathbf{x}', y'; \phi + \delta \phi)  =& \ell_{\text{T}}(\mathbf{x}', y'; \phi) + \mathcal{O}(\varepsilon^2) \\ 
    +& \varepsilon \frac{\nabla_{\phi} \ell_{\text{T}}(\mathbf{x}', y'; \phi) \cdot \nabla_{\phi} \ell_{\text{T}}(\mathbf{x}, y; \phi)}{\|\nabla_{\phi} \ell_{\text{T}}(\mathbf{x}, y; \phi)\|^2}.
\end{split}
\end{equation}

Replacing $\mathbf{x}$ with synthetic data $\hat{z}=(\mathbf{x},y)$ and aggregating its influence across the guidance set $\mathcal{D}_{guide}$ leads to:
\begin{equation}
\begin{split}
    \Delta \mathcal{L}_{T}(\hat{z}) =& -\sum_{(\mathbf{x}',y')\in\mathcal{D}_{guide}}
H_{\phi}\bigl(\mathbf{x},y,\mathbf{x}',y'\bigr)\\
=&  \sum_{(\mathbf{x}', y') \in \mathcal{D}_{guide}} \left[ \ell_{T}(\mathbf{x}', y'; \phi)- \ell_{T}(\mathbf{x}', y'; \phi+\delta \phi)\right] \\
    = &-\sum_{(\mathbf{x}', y') \in \mathcal{D}_{guide}} \varepsilon 
    \frac{\nabla_{\phi} \ell(\mathbf{x}', y'; \phi) \cdot \nabla_{\phi} \ell(\hat{z}; \phi)}{\|\nabla_{\phi} \ell(\hat{z}; \phi)\|^2}. 
\end{split}
\end{equation}

% We now observe that the quantity $H_{\phi}(x, y, x', y')$ introduced in Eq.~\eqref{eq:origin} can be equivalently expressed as:
% \begin{equation}
%     H_{\phi}(x, y, x', y')
%     = \varepsilon\,
%     \frac{\nabla_{\phi} \ell(x', y'; \phi) \cdot \nabla_{\phi} \ell(x, y; \phi)}
%          {\|\nabla_{\phi} \ell(x, y; \phi)\|^2},
% \end{equation}
% which captures the directional alignment between the gradients of $(x, y)$ and $(x', y')$ under model parameters $\phi$.

Notice that we can denote the gradient accumulation of the guidance set $\mathcal{D}_{guide}$ by $\mathbf{G}$, we can end up with the following equation:

\begin{equation}
    \mathbf{G} = -\sum_{(\mathbf{x}', y') \in \mathcal{D}_{guide}} \varepsilon 
    \frac{\nabla_{\phi} \ell(\mathbf{x}', y'; \phi)}{\|\nabla_{\phi} \ell(\hat{z}; \phi)\|^2}
\end{equation}

\begin{equation}
    \Delta \mathcal{L}_{T}(\hat{z}) =   \nabla_{\phi}  \ell(\hat{z}; \phi)\cdot{G}
\end{equation}
 
Where $\Delta \mathcal{L}_{T}(\hat{z})$ measures the positive impact of the synthetic sample $\hat{z}$ on the target loss. Specifically, a greater increase in $\Delta \mathcal{L}_{T}(\hat{z})$ indicates that the model has better optimized its objective function, implying improved generalization to unseen data. This improvement manifests as enhanced performance metrics (e.g., accuracy, AUC, or F1-score) on downstream tasks. By generating synthetic samples that maximize this influence—i.e., by  maximizing $\Delta \mathcal{L}_{T}(\hat{z})$—we aim to augment the dataset with medical time series data that is beneficial for downstream tasks. This process addresses challenges such as data sparsity, class imbalance, and noisy measurements often encountered in medical datasets. By introducing synthetic samples that capture task-relevant patterns (e.g., subtle physiological changes or rare but critical events), the model can learn more representative and clinically significant features, ultimately improving its robustness and reliability in tasks such as mortality prediction and disease diagnosis.

\subsection{TarDiff Pipeline}
\label{sec:igd_medts}

\noindent
In this section, we provide an end-to-end overview of the Influence Guided Diffusion pipeline for generating high-quality medical time-series data customized to a target task~$\mathcal{T}$. The complete procedure is illustrated in Algorithm~\ref{alg:igd}, which consists of three main steps: (1) pre-training the downstream model, (2) computing influence gradients, and (3) performing guided diffusion sampling.

\textbf{Step 1: Pre-train Downstream Model.} We first optimize the downstream task model $f_{\phi}$ on the original dataset $\mathcal{D}_{train}$, aiming to find parameters $\phi^*$ that minimize the target loss function $\ell(\cdot; \phi)$ over $(\mathbf{x},y) \in \mathcal{D}_{train}$. This yields a well-trained model capable of capturing task-specific knowledge relevant to the subsequent generation process.

\textbf{Step 2: Compute Data Influence.} Using the trained parameters $\phi^*$, we then compute per-sample gradients $\nabla_{\phi} \ell(\mathbf{x}_i, y_i; \phi^*)$ for each sample in guidance dataset $\mathcal{D}_{guide}$. Accumulating and normalizing these gradients produces a single vector $\mathbf{G}$ that reflects the aggregated influence of the dataset on the downstream model. This gradient cache $\mathbf{G}$ is leveraged to guide the diffusion model, ensuring that generated samples maximize their impact on the target task.

\textbf{Step 3: Influence-Guided Diffusion Sampling.}  
With $\mathbf{G}$ fixed, we initialize $\mathbf{x}_T$ from a standard Gaussian. At each reverse diffusion step $t$, the model outputs $\mu_t = \mu_\theta(\mathbf{x}_t, y, t)$. We then compute an influence-driven guidance term,  
$\mathbf{J} \gets \nabla_{\mathbf{x}_t} \bigl( \mathbf{G} \cdot \nabla_{\phi} \ell(\mathbf{x}_t, y; \phi^*) \bigr)$,  
 where $\ell$ is the downstream loss, and $\mathbf{G}$ encapsulates the influence from the guidance set. We update the mean via  
$\tilde{\mu}_t \gets \mu_t + w \cdot \mathbf{J}$,  
where $w$ controls the strength of task-oriented guidance. Finally, we sample $\mathbf{x}_{t-1} \sim \mathcal{N}(\tilde{\mu}_t, \Sigma_\theta(t))$. Iterating this procedure through all diffusion steps yields the final synthetic sample $\hat{z} = \mathbf{x}_0$, biased toward improving downstream performance.

By combining a pre-trained downstream task model, influence gradient aggregation, and guidance-based diffusion sampling, TarDiff provides a unified pipeline for generating task-specific synthetic data. This pipeline ensures the fidelity of generated medical time-series data while aligning it closely with the optimization objective of the target task, making it particularly suitable in scenarios such as diagnostic improvement or risk prediction in healthcare applications.

\begin{algorithm}
\caption{TarDiff Pipeline}
\label{alg:igd}
\begin{algorithmic}[H]
\REQUIRE 
    Original dataset $\mathcal{D}_{train}$, guidance subset $\mathcal{D}_{guide}$
    Pretrained conditional diffusion model $\mu_\theta(\mathbf{x}_t,y,t)$,$\Sigma_\theta(t)$. 
    Downstream task model $f_{\phi}$ with random initialization,
    Loss function $\ell(\cdot; \phi)$,
    Total diffusion steps $T$,
    Influence scaling factor $w$.
\ENSURE Synthetic sample $\hat{z}$ optimizing task $\mathcal{T}$.
\STATE \textbf{Step 1: Pre-train downstream model} \\
Optimize $\phi^* \gets \mathop{\mathrm{arg\,min}}_{\phi} \sum_{(\mathbf{x},y) \in \mathcal{D}_{train}} \ell(\mathbf{x}, y; \phi)$
\STATE \textbf{Step 2: Compute Data Influence.} \\
Initialize $\mathbf{G} \gets \mathbf{0}$
\FOR{$i = 1$ to $|{\mathcal{D}}_{guide}|$} 
    \STATE Get sample $(\mathbf{x}_i, y_i) \in \mathcal{D}_{guide}$
    \STATE Compute per-sample gradient: \\
    $\mathbf{g}_i \gets \nabla_{\phi} \ell(\mathbf{x}_i, y_i; \phi^*)$
    \STATE Accumulate gradients: 
    $\mathbf{G} \gets \mathbf{G} + \mathbf{g}_i$
\ENDFOR
\STATE Normalize: 
$\mathbf{G} \gets \frac{1}{|\mathcal{D}_0|} \mathbf{G}$ 
\STATE \textbf{Step 3: Influence Guided diffusion sampling} \\
Initialize $\mathbf{x}_T \sim \mathcal{N}(0, I)$ \\
\FOR{$t = T$ to $1$}
  \STATE (a) $\mu_t \gets \mu_\theta(\mathbf{x}_t, y, t)$.
  \STATE (b) $\mathbf{J} \gets \nabla_{\mathbf{x}_t}\bigl[\mathbf{G} \cdot \nabla_{\phi}\ell(\mathbf{x}_t,y_t; \phi^*)\bigr]$
  \STATE (c) $\tilde{\mu}_t \gets \mu_t + w \cdot \mathbf{J}$.
  \STATE (d) $\mathbf{x}_{t-1} \sim \mathcal{N}(\tilde{\mu}_t, \Sigma_\theta(t))$.
\ENDFOR
\STATE Return $\hat{z} \gets (\mathbf{x}_0,y_0)$\\
\end{algorithmic}
\end{algorithm}

%% file: section/Experiment.tex
\section{Experiment Setup and Result Analysis}

Our experiments are designed to systematically evaluate the proposed method in terms of data quality, augmentation effectiveness, influence guidance mechanism, and computational efficiency. Specifically, we investigate: (1) the feasibility of entirely replacing real data with synthetic data (Section~\ref{tstr}), (2) the effectiveness of synthetic data as augmentation (Section~\ref{tsrtr}), (3) the specific impact of the influence guidance mechanism (Section~\ref{influence_exp}), and (4) the computational efficiency of our proposed method (Section~\ref{sec:complexity}). Throughout all experiments, we adopt the standard validation split as the \textbf{i.i.d.\ guidance set} introduced in Section~\ref{pf}.

\subsection{Datasets, Baselines, and Evaluation Metrics}

\textbf{Datasets.} 
We evaluate our method across multiple datasets. \textbf{MIMIC-III}~\citep{johnson2016mimic} includes multivariate ICU data (7 features, 24 steps) from 20,920 samples. \textbf{eICU}~\citep{pollard2018eicu} provides ICU data (3 features, 288 steps after preprocessed) from over 200,000 admissions. Additionally, we test generalizability on four physiological signal datasets: \textit{APAVA}~\citep{apava}, \textit{PTB}~\citep{ptb}, \textit{TDBRAIN}~\citep{tdbrain}, and \textit{ADFD}~\citep{adfd}, covering diverse ECG and EEG signals. All datasets use an 80\%-10\%-10\% split for training, validation, and test sets. Detailed descriptions and preprocessing steps are provided in Appendix~\ref{Appendix:dataset_details}.

\textbf{Baselines.} 
We compare our approach with several state-of-the-art generative methods: 
\textit{TimeGAN}~\citep{timegan}, a GAN-based model; 
\textit{TimeVAE}~\citep{timevae}, a variational autoencoder model; 
\textit{Diffusion-TS}~\citep{diffusionts}, an unconditional diffusion model; 
% \textit{HaLO}~\citep{halo}, a transformer based model;
\textit{TimeVQVAE}~\citep{timevqvae} and \textit{BioDiffusion}~\citep{DBLP:journals/corr/abs-2401-10282}, both conditional generative models capable of directly generating label-conditioned time series. 
For unconditional models (TimeGAN, TimeVAE, Diffusion-TS), we train separate class-specific models. To evaluate the downstream utility of generated data, we use \textit{TimesNet}~\citep{timesnet}, a state-of-the-art time-series classification architecture, to measure classification performance.

\textbf{Evaluation Metrics.} 
Given the inherent class imbalance in clinical datasets, we use Area Under the Receiver Operating Characteristic Curve (AUROC) and Area Under the Precision-Recall Curve (AUPRC) as primary evaluation metrics. Detailed metric definitions and formulas are provided in Appendix~\ref{Appendix:evaluation_metrics}.

\subsection{Train on Synthetic, Test on Real (TSTR)}
\label{tstr}

 We evaluate the potential of the generated data to serve as a substitute for original data in training high-performance models for clinical tasks. The downstream classifier (\texttt{TimesNet} \citep{timesnet}) is trained exclusively on synthetic data produced by each time-series generation method, then evaluated on real test data to assess its generalization capability.

Table~\ref{tab:performance_comparison1} summarizes the TSTR performance on MIMIC-III and eICU datasets, while the results on high-frequency EEG (\textit{APAVA}, \textit{ADFTD}, \textit{TDBrain}) and ECG (\textit{PTB}) datasets are presented separately in Table~\ref{tab:performance_comparison2}. Consistently, \texttt{TarDiff} achieves state-of-the-art performance in terms of AUROC and AUPRC. Notably, in the TSTR setting, models trained solely on \texttt{TarDiff}-generated samples outperform those trained on synthetic data from other baselines across all tasks, including both standard EHR classification  and EEG/ECG-based diagnoses. Despite having no access to original data during training, the downstream classifiers attain high scores on real test sets, confirming that \texttt{TarDiff}-generated samples effectively capture the essential clinical or physiological features needed for robust predictive modeling.

These findings emphasize that our approach not only produces realistic time-series data but also generates samples that are carefully optimized to improve downstream clinical predictions in a variety of scenarios, thereby validating the robustness and utility of \texttt{TarDiff} in critical healthcare applications ranging from general EHR to high-frequency EEG/ECG analytics.

\begin{table}
    \centering
    \caption{Performance Comparison of Synthetic Data Generation Methods on MIMICIII and eICU}
    \label{tab:performance_comparison1}
    \resizebox{\linewidth}{!}{%

    \begin{tabular}{lcccccccc}
        \toprule
        & \multicolumn{4}{c}{\textbf{MIMIC-III}} & \multicolumn{4}{c}{\textbf{eICU}} \\
        \cmidrule(lr){2-5} \cmidrule(lr){6-9} 
        \textbf{Method} & \multicolumn{2}{c}{\textbf{Mortality}} & \multicolumn{2}{c}{\textbf{ICU Stay}} & \multicolumn{2}{c}{\textbf{Mortality}} & \multicolumn{2}{c}{\textbf{ICU Stay}} \\
        \cmidrule(lr){2-3} \cmidrule(lr){4-5} \cmidrule(lr){6-7} \cmidrule(lr){8-9}
        & \textbf{AUPRC} & \textbf{AUROC} & \textbf{AUPRC} & \textbf{AUROC} & \textbf{AUPRC} & \textbf{AUROC} & \textbf{AUPRC} & \textbf{AUROC} \\
        \midrule
        TimeGAN          & 0.1402 & 0.5144 & 0.3645 & 0.5431 & 0.1592 & 0.6238 & 0.4213 & 0.4620\\
        TimeVAE          & 0.0957 & 0.5392 & 0.3939 & 0.5656 & 0.1046 & 0.5233 & 0.4753 & 0.5321 \\
        TimeVQVAE        & 0.0874 & 0.5182 & 0.3790 & 0.5389 & 0.1216 & 0.5721 & 0.4520 & 0.5287 \\
        % HaLO & 0.0934 & 0.5260 & 0.3770 & 0.5445 & 0.1078 & 0.5188 & 0.4714 & 0.5338 \\

        DiffusionTS      & 0.0865 & 0.5330 & 0.3451 & 0.4946 & 0.1292 & 0.5594 & 0.4692 & 0.5261\\
        BioDiffusion     & 0.0964 & 0.5335 & 0.3370 & 0.4905 & 0.1435 & 0.5872 & 0.4625 & 0.5323\\
         \midrule
         Real Data        & 0.1736 & 0.6350 & 0.4618 & 0.6282 & 0.2072 & 0.6869 & 0.6004 & 0.6615 \\
       \midrule
       TarDiff  & 0.1799 & 0.6373 & 0.4183 & 0.5800 & 0.1698 & 0.6308 & 0.5583 & 0.6184\\
        \bottomrule
    \end{tabular}
}
\end{table}

\begin{table}
    \centering
    \caption{Performance Comparison of Synthetic Data Generation Methods on ECG and EEG Datasets}
    \label{tab:performance_comparison2}
    \resizebox{\linewidth}{!}{%
    \begin{tabular}{lcccccccc}
        \toprule
        \textbf{Method} & \multicolumn{2}{c}{\textbf{APAVA}} & \multicolumn{2}{c}{\textbf{ADFD}} & \multicolumn{2}{c}{\textbf{PTB}} & \multicolumn{2}{c}{\textbf{TDBrain}} \\
        \cmidrule(lr){2-3} \cmidrule(lr){4-5} \cmidrule(lr){6-7} \cmidrule(lr){8-9}
        & \textbf{AUPRC} & \textbf{AUROC} & \textbf{AUPRC} & \textbf{AUROC} & \textbf{AUPRC} & \textbf{AUROC} & \textbf{AUPRC} & \textbf{AUROC} \\
        \midrule
        TimeGAN        & 0.61229          & 0.51033          & 0.35839          & 0.54113          & 0.86766          & 0.78173          & 0.60932 & 0.62780 \\
        TimeVAE        & 0.74266          & 0.68569          & 0.42466          & 0.60053          & 0.95092          & 0.89445          & 0.58464 & 0.58565 \\
        TimeVQVAE      & 0.63722          & 0.55500          & 0.33884          & 0.50578          & 0.94862          & 0.89843          & 0.51336 & 0.54659 \\
        DiffusionTS    & 0.57083          & 0.47062          & 0.33286          & 0.50039          & 0.87372          & 0.75979          & 0.49900 & 0.49319 \\
        BioDiffusion        & 0.63294          & 0.54325          & 0.46042          & 0.63753          & 0.85887          & 0.74805         & 0.55134 & 0.56796 \\
       \midrule

        Real Data & 0.76692 & 0.72063 & 0.43349 & 0.62390 & 0.96768 & 0.93306 & 0.96424 & 0.96153 \\
       \midrule

        TarDiff           & 0.76519 & 0.77097 & 0.47950 & 0.64429 & 0.95435 & 0.90532 & 0.65420 & 0.64444 \\
        \bottomrule
    \end{tabular}
    }
\end{table}

\subsection{Train on Synthetic and Real, Test on Real (TSRTR)}
\label{tsrtr}

\textbf{Setups.} In this experiment, we investigate the effect of incorporating synthetic EHR data—generated by different time series generation framework—into the training set alongside real data, and subsequently testing the trained model on a real test set. We examine five synthetic-to-real mixing ratios: 0.2, 0.4, 0.6, 0.8, and 1.0, where the ratio indicates the size of the synthetic dataset relative to the real training set. For each ratio \(\alpha\), the combined training set is defined as
\[
\mathcal{D}_{\text{train}} = \mathcal{D}_{\text{real}} \cup \mathcal{D}_{\text{synthetic}}(\alpha),
\]
with \(\mathcal{D}_{\text{synthetic}}(\alpha)\) representing the synthetic data sampled to match the specified proportion \(\alpha\).

\begin{figure*}[h]
    \centering
    \includegraphics[width=0.9\textwidth]{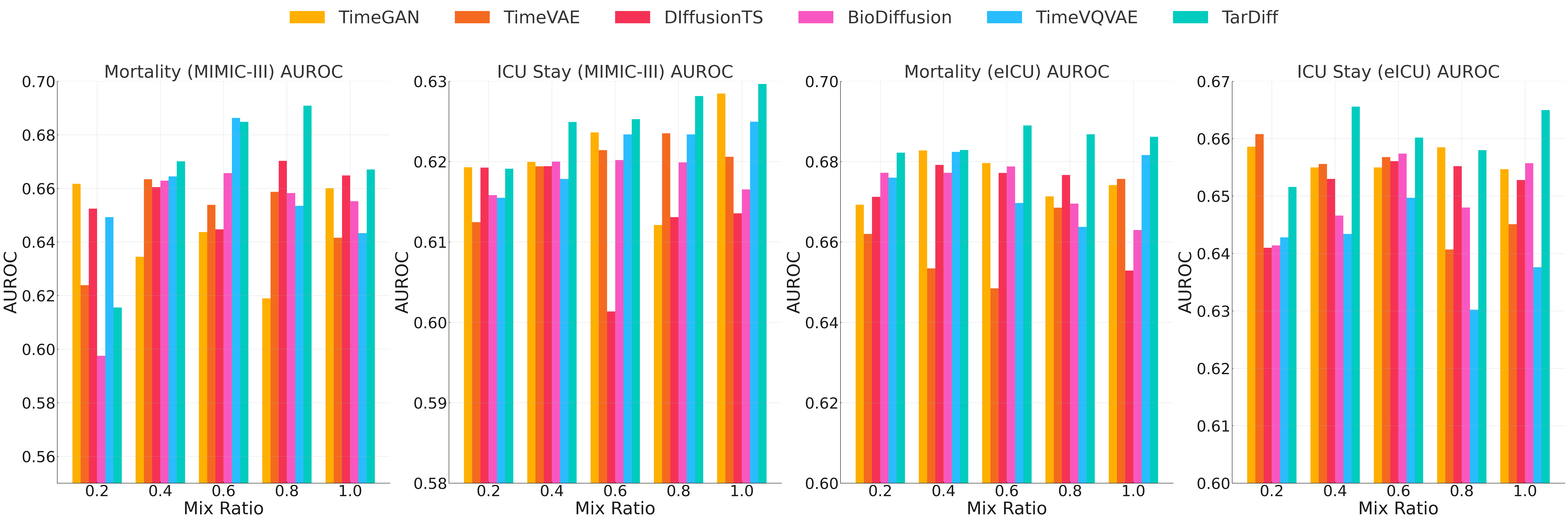}
    \caption{Comparison of AUROC values for the Mortality and ICU Stay task on the MIMIC III and eICU dataset with synthetic-to-real data mix ratios from 0.2 to 1.0.}
    \label{fig:tsrtr}
\end{figure*}

We evaluate the downstream performance on six datasets using key metrics AUROC. Figure~\ref{fig:tsrtr} illustrates the AUROC performance of various methods across two datasets under different synthetic-to-real data mix ratios. Our method demonstrates superior performance, especially at higher mix ratios.

\noindent \textbf{Results.} 
The results indicate that TarDiff outperforms other time series generation baselines across the majority of tasks and mixing ratios, demonstrating a generally upward trend as the synthetic proportion increases from 0.2 to 1.0. Although minor fluctuations occur, TarDiff-derived samples exhibit a sustained positive impact on model performance by leveraging real data guidance throughout the generation process. This synergy enables the synthetic data to capture clinically relevant features more effectively, thereby improving outcomes on these datasets. In contrast, while some baselines occasionally show improvements, their performance tends to degrade or become inconsistent at higher mixing ratios, suggesting a limited capacity to maintain useful signal in purely synthetic settings. Overall, these findings highlight TarDiff’s robust ability to generate synthetic samples that enhance downstream performance and reinforce clinically meaningful patterns when integrated with real data.

Overall, these findings underscore the potential of leveraging synthetic EHR time series data to supplement limited real-world datasets, thereby boosting performance on critical clinical tasks.

\subsection{Influence Guidance under Class Imbalance}\label{sec:imbalance}

This subsection investigates whether influence-guided diffusion mitigates the inherent label imbalance in clinical prediction tasks on MIMIC-III and eICU.

\paragraph{Gradient Analysis.}
We first measure $\ell_{2}$-norms of gradients obtained from a pretrained TimesNet mortality classifier.  
\autoref{tab:grad_norm} indicates that minority samples ($\approx$9–11\% of instances) exhibit substantially larger gradient magnitudes than majority samples, suggesting these cases are intrinsically harder to classify and thus receive stronger guidance signals.

\begin{table}[h]
    \centering
    \caption{Mean gradient norms ($\pm$\,std) for majority vs.\ minority samples.}
    \label{tab:grad_norm}
    \begin{tabular}{lcc}
        \toprule
        \textbf{Dataset} & \textbf{Majority} & \textbf{Minority} \\
        \midrule
        MIMIC-III & $1.06 \pm 1.23$ & $\mathbf{16.85 \pm 2.48}$ \\
        eICU      & $5.41 \pm 5.05$ & $\mathbf{37.86 \pm 8.73}$ \\
        \bottomrule
    \end{tabular}
\end{table}

\paragraph{Minority-Class Performance.}
\autoref{tab:minority_f1} reports minority-class F$_1$ scores when real data are augmented with synthetic data generated by TarDiff (TSRTR protocol).  
Compared with a real-only baseline, TarDiff more than doubles the minority F$_1$ on MIMIC-III (+93\%) and improves eICU by 44\%, demonstrating that unified influence guidance already alleviates imbalance without explicit class weighting.

\begin{table}[h]
    \centering
    \caption{Minority-class F$_1$ under different generation strategies.}
    \label{tab:minority_f1}
    \begin{tabular}{lcc}
        \toprule
        \textbf{Method} & \textbf{MIMIC-III} & \textbf{eICU} \\
        \midrule
        TRTR (Real-only) & 0.056 & 0.013 \\
        TarDiff          & \textbf{0.108} & \textbf{0.018} \\
        \bottomrule
    \end{tabular}
\end{table}

\paragraph{Class-Specific Guidance.}
To further isolate the effect of guidance signals, gradients are recomputed on (i) all guidance samples, (ii) majority-only samples, and (iii) minority-only samples.  
\autoref{tab:class_guidance} shows that minority-only guidance attains the highest minority F$_1$ (0.163 on MIMIC-III; 0.025 on eICU), while majority-only guidance degrades performance.

\begin{table}[h]
    \centering
    \caption{Minority-class F$_1$ with class-specific guidance.}
    \label{tab:class_guidance}
    \begin{tabular}{lcc}
        \toprule
        \textbf{Guidance Source} & \textbf{MIMIC-III} & \textbf{eICU} \\
        \midrule
        All Samples          & 0.108 & 0.018 \\
        Majority-only        & 0.066 & 0.012 \\
        \textbf{Minority-only} & \textbf{0.163} & \textbf{0.025} \\
        \bottomrule
    \end{tabular}
\end{table}

\paragraph{Discussion.}
The pronounced gradient disparity (\autoref{tab:grad_norm}) indicates that influence guidance naturally focuses on under-represented events.  
Consequently, TarDiff improves rare-class metrics without additional hyper-parameters and retains flexibility to apply targeted gradients for further gains, providing a principled mechanism for alleviating class imbalance in medical time-series generation.

\subsection{Complexity Analysis.}
\label{sec:complexity}

We analyze the additional computational costs introduced by our gradient-guided diffusion framework relative to a standard diffusion-based generation pipeline. First, consider the one-time overhead of training a downstream task network (e.g., a classifier or regressor) and caching its gradients over a target set. Specifically, the downstream model must be trained until convergence, followed by forward-backward passes on the target set to obtain gradient norms with respect to the downstream loss. If $N_{\text{t}}$ is the size of the target set, $f_{\text{T}}(L, D)$ represents the complexity of a single forward-backward pass for the downstream model on time-series data of length $L$ and feature dimensionality $D$, and if these gradients are stored once for reuse, the overall cost of this stage can be approximated by
\begin{equation}
   O\bigl(N_{\text{t}} \cdot f_{\text{T}}(L, D)\bigr).
\end{equation}
In practice, on eight real-world datasets, we measure this one-time overhead to range from 10s to 167s; please see the Appendix \ref{Appendix:Runtime_Comparison} for the detailed statistics.

Next, during sampling, we incorporate gradient guidance by projecting intermediate samples $\mathbf{x}_t$ onto directions derived from the cached gradient norms. Because the downstream network is not re-invoked at each diffusion step, the extra per-step overhead is limited to dot products and final gradient computations, denoted by $g(L, D)$. Hence, for $T$ diffusion steps and a batch size $B_{\text{sample}}$, the cost of gradient-guided sampling is
\begin{equation}
   O\bigl(T \cdot B_{\text{sample}} \cdot g(L, D)\bigr).
\end{equation}

In comparison, a standard diffusion framework (e.g., DDPM) typically incurs a sampling cost of 
\begin{equation}
   O\bigl(T \cdot B_{\text{sample}} \cdot h(L, D)\bigr),
\end{equation}
where $h(L, D)$ is the computational complexity for each step without gradient guidance. Consequently, the additional overhead ratio can be approximated by
\begin{equation}
   \frac{T \cdot B_{\text{sample}} \cdot g(L, D)}{T \cdot B_{\text{sample}} \cdot h(L, D)} 
   \;=\; \frac{g(L, D)}{h(L, D)},
\end{equation}
which is typically small because $g(\cdot)$ involves only lightweight vector or matrix operations.
Moreover, we compare our sampling speed with other baselines on a diffusion-based model and observe faster sampling for our gradient-guided approach. Full results can be found in the Appendix \ref{Appendix:Runtime_Comparison}.
Overall, our gradient guidance requires a one-time downstream network training plus negligible extra work at each sampling step, yet yields a substantial improvement in aligning generated samples with downstream tasks—an important advantage in applications such as medical time-series data.

\subsection{Sample Influence with Performance}
\label{influence_exp}

To rigorously assess the effectiveness of our proposed TarDiff, we conduct experiments targeting two objectives: (1) investigate whether TarDiff can successfully modulate the influence $\Delta \mathcal{L}_{T}(\hat{z})$ of generated samples; and (2) evaluate how these influence adjustments affect model performance in downstream classification tasks.

Specifically, we partition the original validation set into two subsets: (i) \textbf{Guidance-Val subset}, used exclusively for generating samples during the guidance diffusion process, and (ii) \textbf{Evaluation-Val subset}, used for selecting the optimal guidance scale by assessing downstream task performance. After selecting the optimal guidance scale, we report the final model performance on the \textbf{entire validation set}, thus ensuring an unbiased assessment of TarDiff's generalization capabilities.

\begin{figure}[h]
    \centering
    \includegraphics[width=\linewidth]{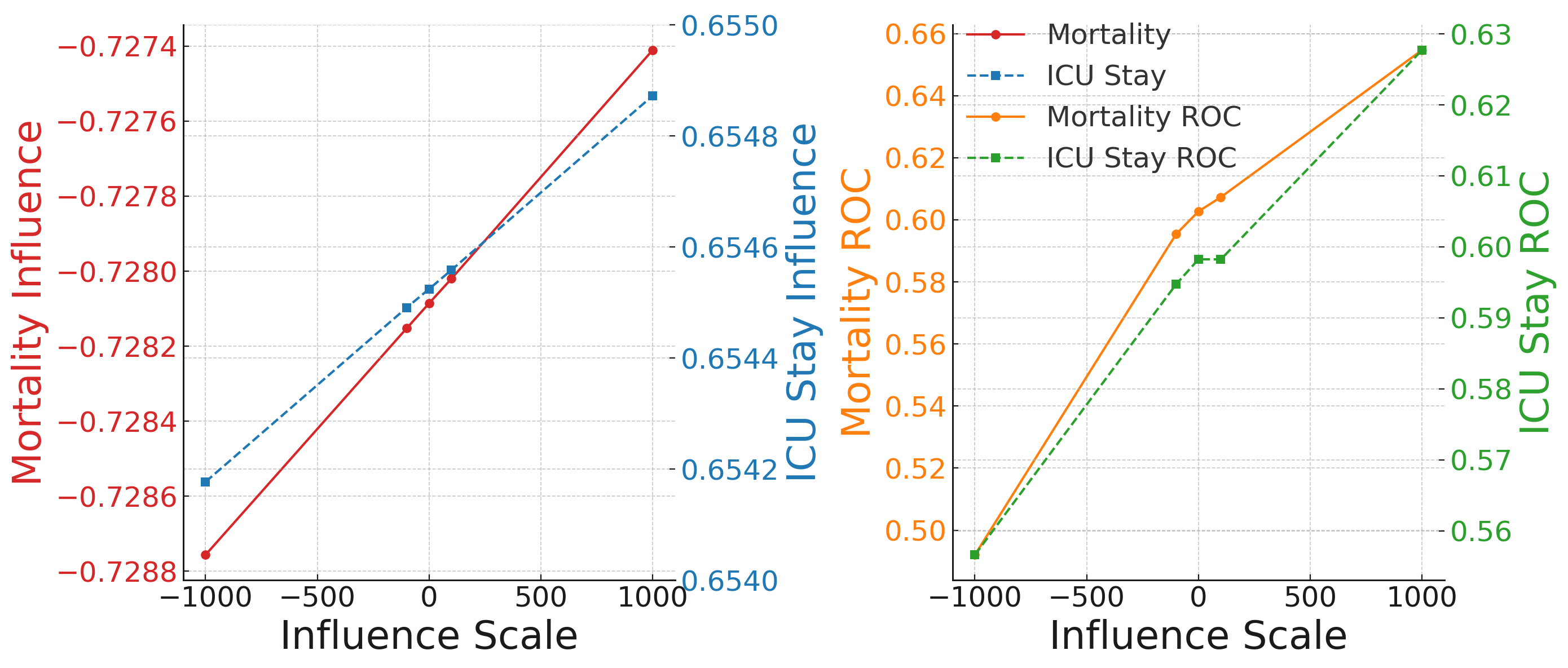}
    \caption{Influence scale analysis conducted by generating samples from the \emph{Guidance-Val subset} and assessing AUROC performance on the \emph{Guidance-Val subset} for Mortality and ICU Stay tasks, with scales ranging from -1000 to 1000. The left panel illustrates sample influence value changes, while the right panel shows AUROC performance across different scales.}
    \label{fig:guidance_scale_effect}
\end{figure}

Figure~\ref{fig:guidance_scale_effect} illustrates experimental outcomes across various influence scales using samples generated from the \emph{Guidance-Val subset}, with performance metrics evaluated on the \emph{Evaluation-Val subset}. The figure comprises two panels: (i) the left panel depicts changes in sample influence values for both Mortality and ICU Stay tasks, and (ii) the right panel reports the corresponding AUROC performance for these tasks.

As shown in the left panel of Figure~\ref{fig:guidance_scale_effect}, varying the influence scale from $-1000$ to $+1000$ markedly affects sample influence values. For the Mortality task, the sample influence consistently declines as the scale increases; conversely, the ICU Stay task exhibits the opposite trend. Correspondingly, in the right panel, these influence adjustments yield improvements in AUROC performance, with both tasks reaching optimal performance at moderate-to-high influence scales (around $+1000$).

By separately tuning the guidance scale using clearly defined subsets of the validation data and subsequently evaluating the final performance on the entire validation set, we ensure the reported results accurately reflect TarDiff’s effectiveness and generalization capabilities.

%% file: section/related_work.tex
\section{Related work}

%\subsection{Healthcare Time Series Generation}
Recent advances in time series generation leverage deep generative models such as GANs, VAEs, and diffusion-based approaches \citep{timegan, timevae, timevqvae, timedp}. TimeGAN \citep{timegan} combines adversarial training with supervised embedding, thereby aligning the generated sequences with the real data’s temporal structure. Meanwhile, TimeVAE \citep{timevae} and TimeVQVAE \citep{timevqvae} adopt latent representations to capture salient patterns, which helps improve both the reconstruction quality and overall fidelity of the synthetic time series. Diffusion-based models like TimeDP \citep{timedp} further enhance realism by incorporating domain prompts in their denoising process, enabling more accurate generation of complex temporal signals. Empirically, these approaches have demonstrated credible temporal fidelity in diverse domains such as finance \citep{yuhaofin} and medicine \citep{12-lead}, leading to promising applications in areas like data augmentation, anomaly detection, and privacy-preserving analytics.

In healthcare applications, generating synthetic patient time series presents unique challenges, including privacy protection and the need for clinically relevant patterns. GAN-based methods have been widely used for realistic EHR synthesis \citep{DBLP:journals/corr/ChoiBMDSS17}, and diffusion models have also emerged as a promising approach, demonstrating efficacy in producing high-fidelity synthetic EHR time-series data \citep{private,DBLP:journals/corr/abs-2402-06318}.  Conditioning on clinical variables, as seen in methods like MEGAN \citep{megan}, enables high-fidelity, multi-perspective ECG generation, thereby facilitating applications like data augmentation and simulation-based scenario testing in clinical research.

%% file: section/conclusion.tex
\section{Conclusion.}
In this paper, we present a task-based framework for electronic medical record time series generation that guides diffusion in generating synthetic data by estimating the impact of synthetic samples on specific downstream task models, maximising the impact of synthetic data on clinical tasks.
Comprehensive experiments on six datasets demonstrate that our framework not only improves the influence of the generated data on the target task. but also significantly enhances downstream model performance.
These results underscore the potential of TarDiff to mitigate data scarcity and privacy concerns in healthcare.

%% file: section/appendix.tex
\section{Dataset Details}
\label{Appendix:dataset_details}
\subsection{Critical Care EHR Datasets (MIMIC-III and eICU)}

\textbf{MIMIC-III}\citep{johnson2016mimic} is a large, publicly available database comprising de-identified health-related data associated with over 40,000 patients who stayed in critical care units of the Beth Israel Deaconess Medical Center between 2001 and 2012 . For our analysis, we focus on the first 24 hours of hospitalization for each patient, resulting in 20,920 samples. Each sample is a 24-step multivariate time series with 7 features: \textit{heart rate}, \textit{systolic blood pressure}, \textit{diastolic blood pressure}, \textit{mean blood pressure}, \textit{respiration rate}, \textit{temperature}, and \textit{oxygen saturation}. We split this dataset into training, validation, and test sets with proportions of 80\%, 10\%, and 10\%, respectively.

\textbf{eICU}\citep{pollard2018eicu} is a multi-center critical care dataset comprising de-identified health data from over 200,000 admissions to intensive care units (ICUs) across the United States between 2014 and 2015. For our analysis, we extract time-series measurements (\textit{heart rate}, \textit{respiratory rate}, and \textit{oxygen saturation}) from the initial 24-hour window of ICU admission. Data are sampled every 5 minutes, resulting in 288 time steps. Each time step includes 3 features, providing a granular view of patient status. The dataset is partitioned into training, validation, and test sets using an 80\%, 10\%, 10\% ratio.

\textbf{Tasks.}   %可以单独写一个，附录放evaluation的公式
We evaluate our framework on two tasks: \emph{(i)} \textit{Mortality Prediction:} Determine whether a patient will die during the hospital stay. In the MIMIC-III dataset, the positive-to-negative ratio is \(1{,}680:19{,}240\), while in the eICU dataset, it is \(3{,}173:27{,}892\). \emph{(ii)} \textit{ICU Length-of-Stay Prediction:} Predict whether a patient's ICU stay exceeds three days. For the MIMIC-III dataset, the positive-to-negative ratio is \(2{,}869:18{,}051\), and for the eICU dataset, it is \(13{,}206:17{,}859\).

Table~\ref{tab:label_distribution} provides detailed information on training label distributions within each dataset. Given the inherent label imbalance in both MIMIC-III and eICU, generating clinically meaningful synthetic data remains a significant challenge.

% \begin{table*}[ht]
%     \centering
%     \caption{MIMIC-III and eICU Dataset Overview.}
%     \label{tab:label_distribution}
%     \begin{tabular}{llccccc}
%         \toprule
%         \textbf{Dataset} & \textbf{Task} & \textbf{All Samples} & \textbf{Negative Samples} & \textbf{Positive Samples} & \textbf{Features} & \textbf{Sequence Length} \\
%         \midrule
%         MIMIC  & Mortality     & \textbf{20,920} & 19,240 & 1,680 & 7 & 24 \\
%             & ICU Stay      & \textbf{20,920} & 18,051 & 2,869 & 7 & 24 \\
%         \midrule
%         eICU   & Mortality     & \textbf{31,065} & 27,892 & 3,173 & 3 & 288 \\
%            & ICU Stay      & \textbf{31,065} & 17,859 & 13,206 & 3 & 288 \\
%         \bottomrule
%     \end{tabular}
% \end{table*}

\begin{table*}[ht]
    \centering
    \caption{MIMIC-III and eICU Dataset Overview.}
    \label{tab:label_distribution}
    \begin{adjustbox}{width=\textwidth}
    \begin{tabular}{llccccc}
        \toprule
        \textbf{Dataset} & \textbf{Task} & \textbf{All Samples} & \textbf{Negative Samples} & \textbf{Positive Samples} & \textbf{Features} & \textbf{Seq Length} \\
        \midrule
        MIMIC  & Mortality     & \textbf{20,920} & 19,240 & 1,680 & 7 & 24 \\
               & ICU Stay      & \textbf{20,920} & 18,051 & 2,869 & 7 & 24 \\
        \midrule
        eICU   & Mortality     & \textbf{31,065} & 27,892 & 3,173 & 3 & 288 \\
               & ICU Stay      & \textbf{31,065} & 17,859 & 13,206 & 3 & 288 \\
        \bottomrule
    \end{tabular}
    \end{adjustbox}
\end{table*}

\subsection{Specialized Physiological Signal Datasets (EEG and ECG)}
For these four datasets, we followed the preprocessing and data partitioning settings (training, validation, and test splits) described in \citep{medformer}. Below, we provide a brief introduction to each dataset and their associated tasks.

\textbf{APAVA} dataset is a public EEG time series dataset comprising recordings from 23 subjects, including 12 Alzheimer's disease (AD) patients and 11 healthy controls. Each subject underwent approximately 30 trials, with each trial consisting of a 5-second EEG recording sampled at 256Hz across 16 channels. The task associated with APAVA is binary classification, distinguishing AD patients from healthy individuals.

\textbf{ADFTD} (Alzheimer's Disease and Frontotemporal Dementia) dataset is an EEG dataset specifically curated to study Alzheimer's disease (AD) and frontotemporal dementia (FTD). It includes EEG recordings from subjects diagnosed with AD, subjects diagnosed with FTD, and healthy controls. EEG signals are recorded across multiple channels at approximately 500Hz and undergo standard preprocessing steps such as filtering and downsampling. The task for ADFTD is a three-class classification to differentiate among AD, FTD, and healthy subjects.

\textbf{PTB} Diagnostic ECG Database is a publicly available collection of 549 high-resolution 15-lead ECG recordings from 290 subjects, aged between 17 and 87 years. Each recording includes the standard 12 leads along with 3 Frank leads (Vx, Vy, Vz), digitized at 1000Hz with 16-bit resolution over a ±16.384 mV range. The database encompasses a variety of cardiac conditions, including myocardial infarction, cardiomyopathy, and bundle branch block, as well as recordings from healthy controls. The task on the PTB dataset is binary classification between patients diagnosed with myocardial infarction and healthy controls.

\textbf{TDBrain} dataset comprises EEG recordings from multiple channels collected from subjects performing eye-closed tasks. The dataset includes EEG data from subjects diagnosed with Parkinson's disease and healthy controls. The associated task is binary classification to distinguish Parkinson's disease patients from healthy individuals.

\subsection{Scalability Across Different Dataset Sizes}

We conducted experiments across multiple datasets varying significantly in sample size, channel numbers, and sequence lengths. This diversity enables us to evaluate the scalability and robustness of our model comprehensively. The datasets cover diverse medical signals, including EHR (MIMIC-III, eICU), ECG (PTB), and EEG (ADFD, APAVA, TDBRAIN), with varying complexity and dimensionality.

As summarized in Table~\ref{tab:dataset_summary}, our method consistently demonstrates solid performance improvements across datasets of different sizes and modalities.

\begin{table}[h]
\centering
\caption{Dataset scales summary}
\label{tab:dataset_summary}
\begin{tabular}{lcccc}
\toprule
\textbf{Dataset} & \textbf{Samples} & \textbf{Channels} & \textbf{Length}  \\ 
\midrule
eICU             & 31,065           & 3                 & 288             \\
MIMIC-III        & 26,150           & 7                 & 24              \\
ADFD             & 69,752           & 19                & 256             \\
PTB              & 64,356           & 15                & 288             \\
TDBRAIN          & 6,240            & 33                & 256             \\
APAVA            & 5,967            & 16                & 256             \\
\bottomrule
\end{tabular}
\end{table}

\subsection{Controllability of Guidance Set Scale}

\begin{figure}[h]
    \centering
    \includegraphics[width=\linewidth]{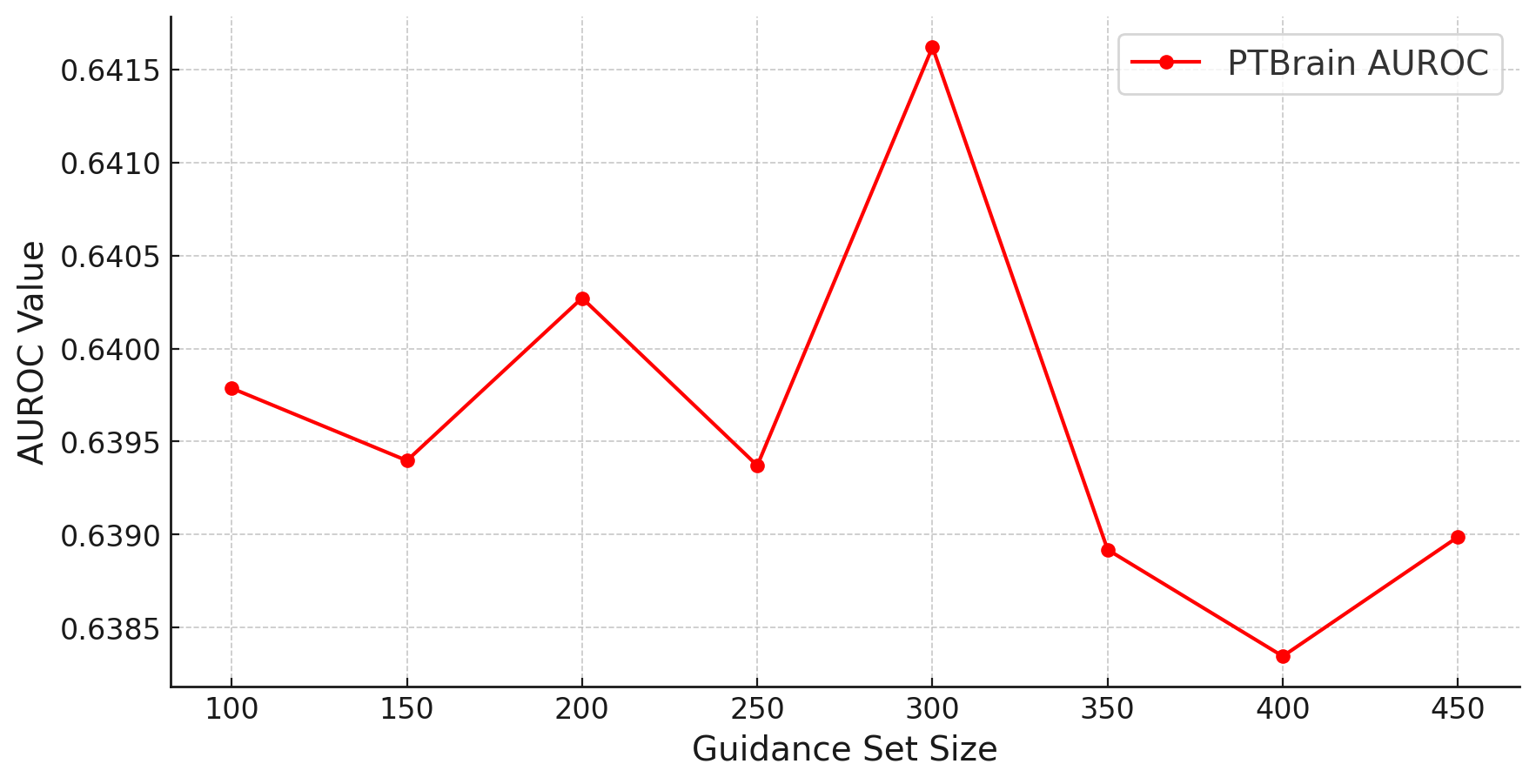}
    \caption{Model performance with varying guidance set sizes on PTBrain dataset.}
    \label{fig:guidance_scale}
\end{figure}

To further demonstrate the scalability and controllability of our guidance mechanism, we conducted an experiment analyzing the impact of varying guidance set sizes on model performance using the PTBrain dataset. As illustrated in Figure~\ref{fig:guidance_scale}, despite fluctuations, the model's performance remains consistently stable across different guidance set sizes. This indicates that our proposed method provides effective control over the guidance scale, allowing users to flexibly adjust it according to practical dataset constraints or computational resources. Consequently, this mitigates potential concerns regarding the scalability and practical applicability of our approach.

\section{Evaluation Metrics Details}\label{Appendix:evaluation_metrics}

\textbf{Area Under the Receiver Operating Characteristic Curve (AUROC)} measures the ability of a classifier to distinguish between classes. It is calculated as follows:
\[
\text{AUROC} = \int_{0}^{1} \text{TPR}(x)\, d(\text{FPR}(x))
\]
where TPR is the True Positive Rate and FPR is the False Positive Rate across different decision thresholds.

\textbf{Area Under the Precision-Recall Curve (AUPRC)} evaluates the classifier’s performance in imbalanced classification problems by considering precision and recall:
\[
\text{AUPRC} = \int_{0}^{1} \text{Precision}(x)\, d(\text{Recall}(x))
\]
where precision is the fraction of relevant instances among retrieved instances, and recall is the fraction of relevant instances retrieved over the total relevant instances.

These metrics provide a comprehensive evaluation of classification models, particularly useful in healthcare scenarios with significant class imbalance.

\section{Runtime and Overhead Comparison}
\label{Appendix:Runtime_Comparison}

As briefly discussed in Section~\ref{sec:complexity}, our approach introduces a one-time overhead for training and gradient caching, as well as a per-step overhead during sampling. This appendix details the runtime measurements for both the one-time overhead and the sampling process. 

\subsection{One-time Overhead}
Table~\ref{tab:one_time_overhead} shows the one-time cost for gradient caching on eight different datasets, which ranges from 10s to 167s.

\begin{table}[ht]
    \centering
    \caption{One-time cost for gradient caching.}
    \label{tab:one_time_overhead}
    \begin{tabular}{l c}
    \toprule
    \textbf{Dataset/Task} & \textbf{Overhead (s)} \\
    \midrule
    TDBRAIN           & 10.51 \\
    APAVA             & 15.49 \\
    ADFD              & 167.81 \\
    PTB               & 144.65 \\
    eICU\_mortality   & 34.01 \\
    eICU\_ICUStay         & 33.93 \\
    MIMIC\_mortality  & 27.51 \\
    MIMIC\_ICUStay        & 27.01 \\
    \bottomrule
    \end{tabular}
\end{table}

\subsection{Sampling Runtime Comparison}
To evaluate the sampling efficiency, we compare our TarDiff against representative GAN-based, 
VAE-based, and Diffusion-based approaches. As seen in Table~\ref{tab:runtime_compare}, 
our method achieves competitive or faster sampling compared to other diffusion-based methods.

\begin{table}[ht]
    \centering
    \caption{Sampling runtime comparison across different generation methods.}
    \label{tab:runtime_compare}
    \begin{tabular}{llc}
        \toprule
        \textbf{Backbone Type} & \textbf{Method} & \textbf{Sampling Time (s/sample)} \\
        \midrule
        \multirow{1}{*}{GAN-based} 
         & TimeGAN        & 0.0005 \\
        \midrule
        \multirow{2}{*}{VAE-based} 
         & TimeVQVAE      & 0.0047 \\
         & TimeVQE        & 0.0006 \\
        \midrule
        \multirow{3}{*}{Diffusion-based} 
         & BioDiffusion   & 0.3008 \\
         & DiffusionTS    & 0.1340 \\
         & \textbf{TarDiff}  & \textbf{0.0259} \\
        \bottomrule
    \end{tabular}
\end{table}

\section{Model Structure and Implementation Details}

The denoising network in our diffusion model employs a one-dimensional U-Net architecture specifically designed for multi-channel time-series data. The model initializes with 64 channels and features multiple resolution levels, each comprising three residual blocks. We apply progressive channel multipliers of [1, 2, 4, 4] to enhance feature representation at coarser resolutions. To effectively capture long-range temporal dependencies, attention mechanisms with eight heads are incorporated at resolutions of 1, 2, and 4. Additionally, the model integrates scale-shift normalization and residual connections for up-sampling and down-sampling to stabilize training. Contextual embeddings are projected into a 32-dimensional latent space. Furthermore, the architecture supports classifier-free guidance and spatial transformations, optimizing its performance for classification tasks.

Training was conducted using a batch size of 256 for 20,000 iterations with a fixed learning rate of 0.0001. During sampling, we consistently employed a guidance scale of 100 for generated samples. All experiments were carried out on a single NVIDIA A100 GPU with 80GB of memory.

\section{Additional Results on Fidelity \& Privacy}\label{app:fidelity_privacy}

\begin{table}[h]
    \centering
    \caption{Distribution Similarity (DS) $\downarrow$}
    \label{tab:ds_full}
    \begin{tabular}{lcccc}
        \toprule
        & \multicolumn{2}{c}{\textbf{MIMIC}} & \multicolumn{2}{c}{\textbf{eICU}}\\
        \cmidrule(lr){2-3}\cmidrule(lr){4-5}
        \textbf{Method} & Mortality & ICU\,Stay & ICU\,Stay & Mortality\\
        \midrule
        TarDiff       & 0.000201 & 0.000000 & 0.1488 & 0.1810\\
        TimeGAN       & 0.000201 & 0.000000 & 0.3781 & 0.2471\\
        TimeVAE       & 0.000000 & 0.0304   & 0.3668 & 0.1709\\
        TimeVQ‑VAE    & 0.0325   & 0.0015   & 0.0000 & 0.3663\\
        DiffusionTS   & 0.000101 & 0.4451   & 0.5000 & 0.4931\\
        BioDiffusion  & 0.000000 & 0.4989   & 0.4968 & 0.5000\\
        \bottomrule
    \end{tabular}
\end{table}

\begin{table}[h]
    \centering
    \caption{Membership\,Inference\,Risk (MIR) $\downarrow$}
    \label{tab:mir_full}
    \begin{tabular}{lcccc}
        \toprule
        & \multicolumn{2}{c}{\textbf{MIMIC}} & \multicolumn{2}{c}{\textbf{eICU}}\\
        \cmidrule(lr){2-3}\cmidrule(lr){4-5}
        \textbf{Method} & Mortality & ICU\,Stay & ICU\,Stay & Mortality\\
        \midrule
        TarDiff       & 0.6761 & 0.6787 & 0.6667 & 0.6667\\
        BioDiffusion  & 0.7316 & 0.8114 & 0.7736 & 0.8199\\
        TimeVQ‑VAE    & 0.6792 & 0.6818 & 0.6667 & 0.6668\\
        TimeGAN       & 0.6949 & 0.6762 & 0.6668 & 0.6668\\
        TimeVAE       & 0.6788 & 0.9169 & 0.6667 & 0.6668\\
        DiffusionTS   & 0.9683 & 0.6811 & 0.9349 & 0.9976\\
        \bottomrule
    \end{tabular}
\end{table}

\begin{table}[h]
    \centering
    \caption{Privacy Score (PS) $\downarrow$}
    \label{tab:ps_full}
    \begin{tabular}{lcccc}
        \toprule
        & \multicolumn{2}{c}{\textbf{MIMIC}} & \multicolumn{2}{c}{\textbf{eICU}}\\
        \cmidrule(lr){2-3}\cmidrule(lr){4-5}
        \textbf{Method} & Mortality & ICU\,Stay & ICU\,Stay & Mortality\\
        \midrule
        TarDiff       & 0.5819 & 0.5669 & 0.5795 & 0.5770\\
        BioDiffusion  & 0.6226 & 0.6656 & 0.5954 & 0.5744\\
        TimeVQ‑VAE    & 1.2898 & 1.2115 & 0.5006 & 0.5671\\
        TimeGAN       & 0.7035 & 0.8837 & 0.5198 & 0.6420\\
        TimeVAE       & 0.9856 & 0.9616 & 0.5071 & 0.5028\\
        DiffusionTS   & 0.8671 & 0.9097 & 0.6558 & 0.6821\\
        \bottomrule
    \end{tabular}
\end{table}

To demonstrate TarDiff’s fidelity and privacy preservation, we provide Discriminative Score (DS, measures how easily a classifier can distinguish synthetic samples from real ones.), Predictive Score (PS, evaluates how accurately models trained on synthetic data perform when predicting outcomes on real test data.), and Membership Inference Risk (MIR, assesses the privacy risk by quantifying vulnerability to membership inference attacks.)

\section{Visualization of Synthetic Data}

We visualize the positive and negative samples generated by different methods. The results indicate that while TimeGAN performs relatively well for positive samples, its negative samples exhibit significant fluctuations and lack trend variations. Additionally, TimeVAE and TimeVQVAE produce overly smoothed sequences. In contrast, our approach more closely aligns with the distribution of real data.

\begin{figure*}
    \centering
    \scalebox{0.8}{
    \includegraphics[width=\textwidth]{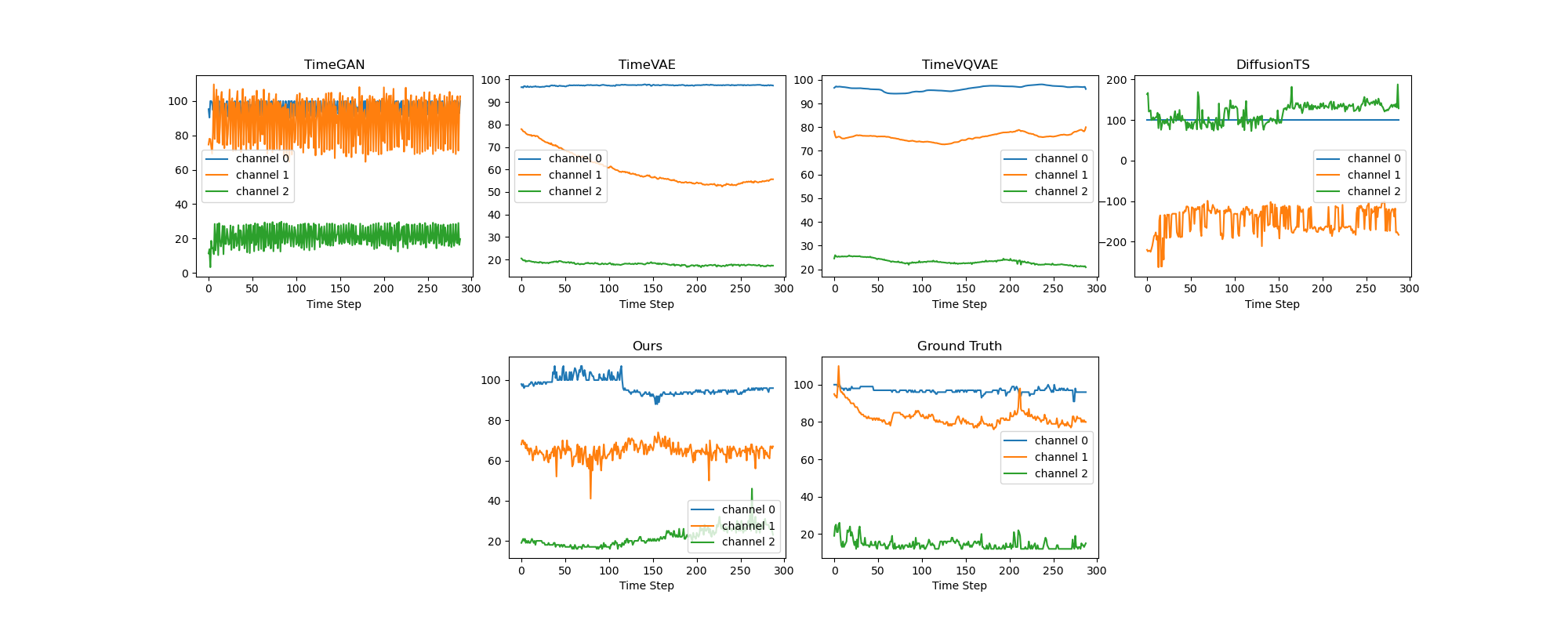}
    }
    \caption{Visualization of negative samples generated by different methods for ICU-Stay on eICU}
    \label{eicu_icustay_neg}
\end{figure*}

\begin{figure*}
    \centering
    \scalebox{0.8}{
    \includegraphics[width=\textwidth]{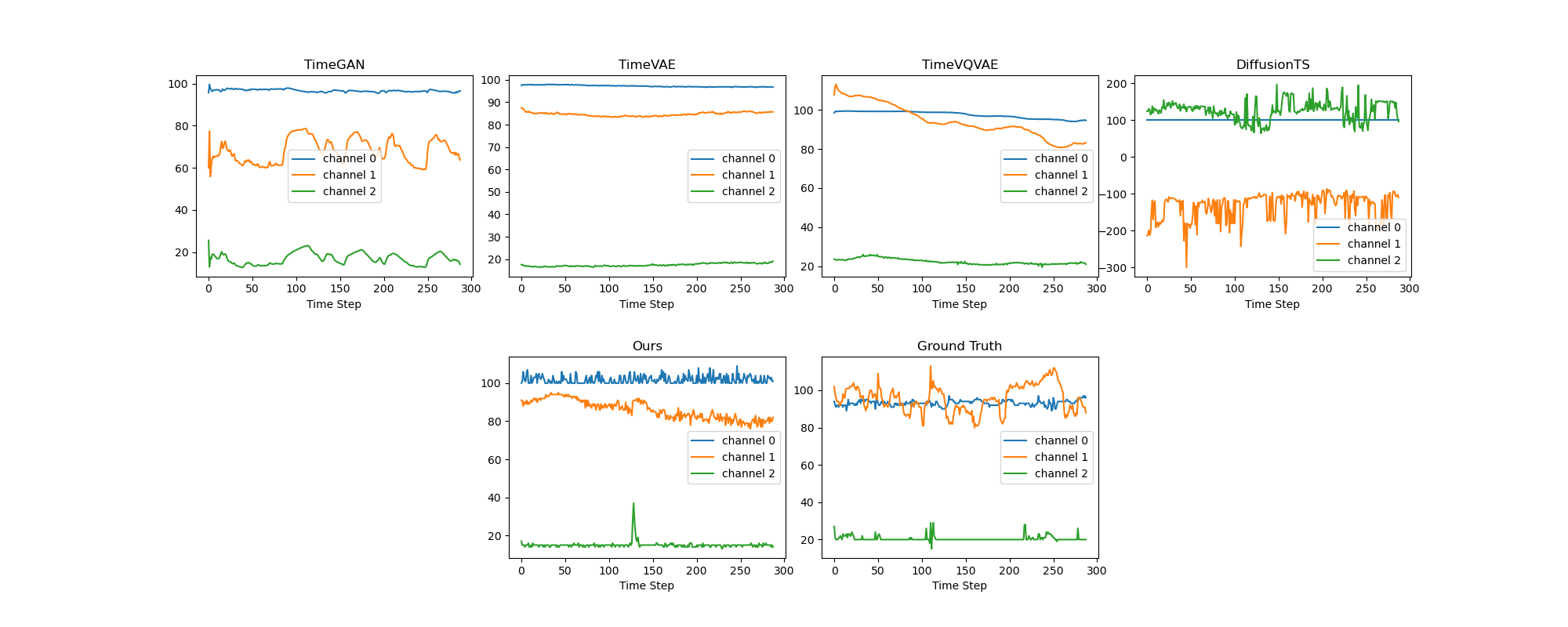}
    }
    \caption{Visualization of positive samples generated by different methods for ICU-Stay on eICU}
    \label{eicu_icustay_pos}
\end{figure*}

\begin{figure*}
    \centering
    \scalebox{0.8}{
    \includegraphics[width=\textwidth]{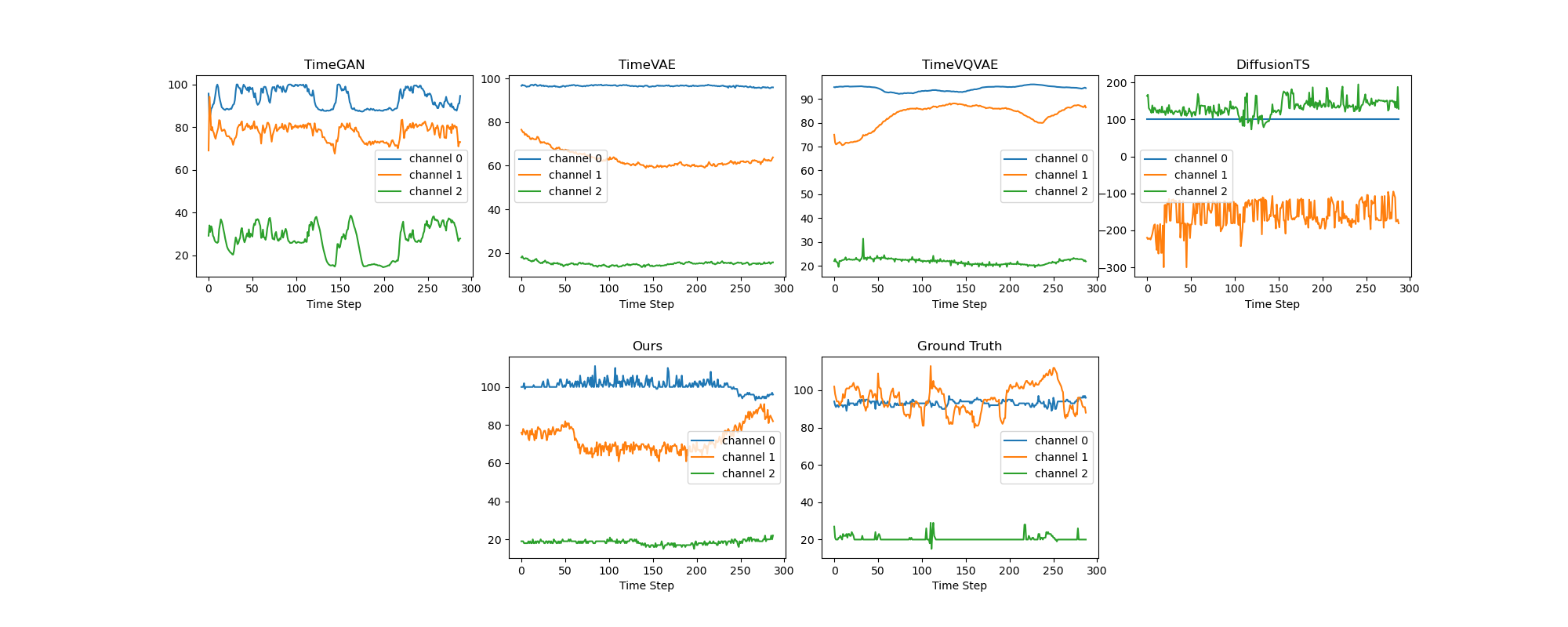}
    }
    \caption{Visualization of negative samples generated by different methods for Mortality on eICU}
    \label{eicu_motality_neg}
\end{figure*}

\begin{figure*}
    \centering
    \scalebox{0.8}{
    \includegraphics[width=\textwidth]{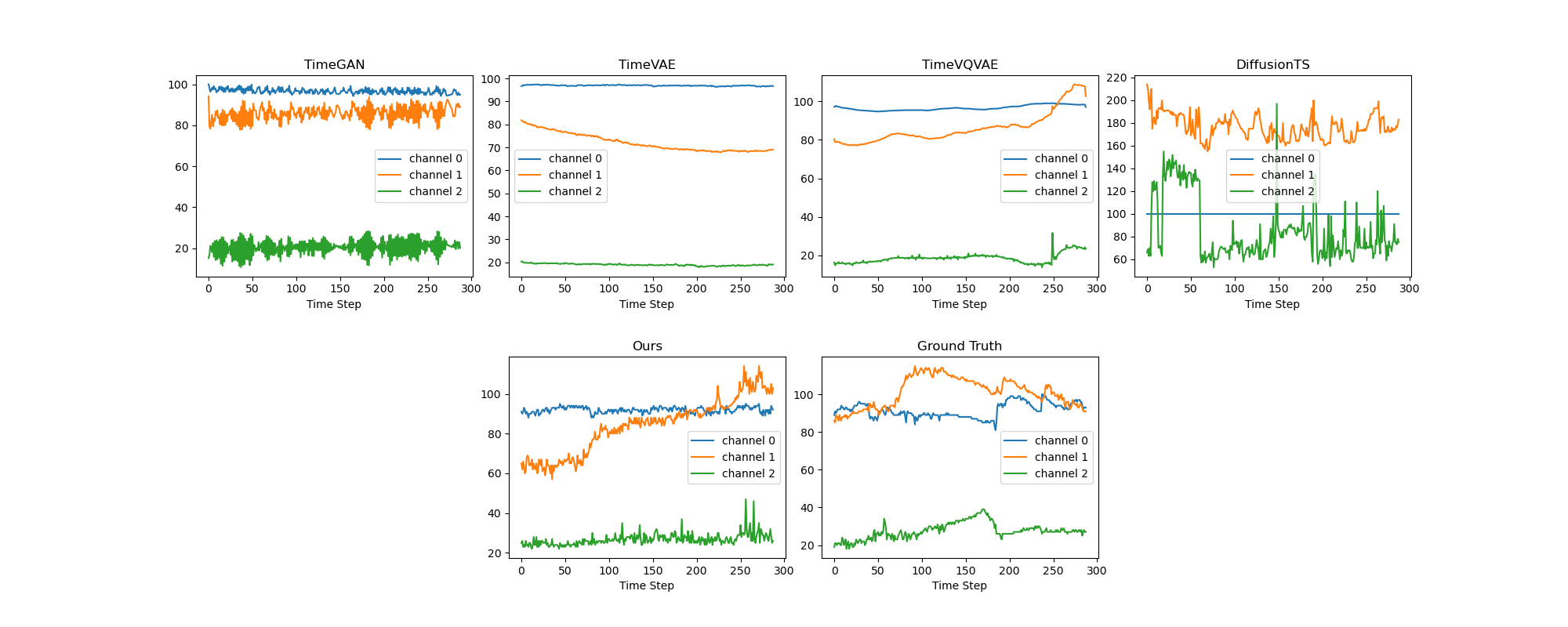}
    }
    \caption{Visualization of positive samples generated by different methods for Mortality on eICU}
    \label{eicu_motality_pos}
\end{figure*}

\begin{figure*}
    \centering
    \scalebox{0.8}{
    \includegraphics[width=\textwidth]{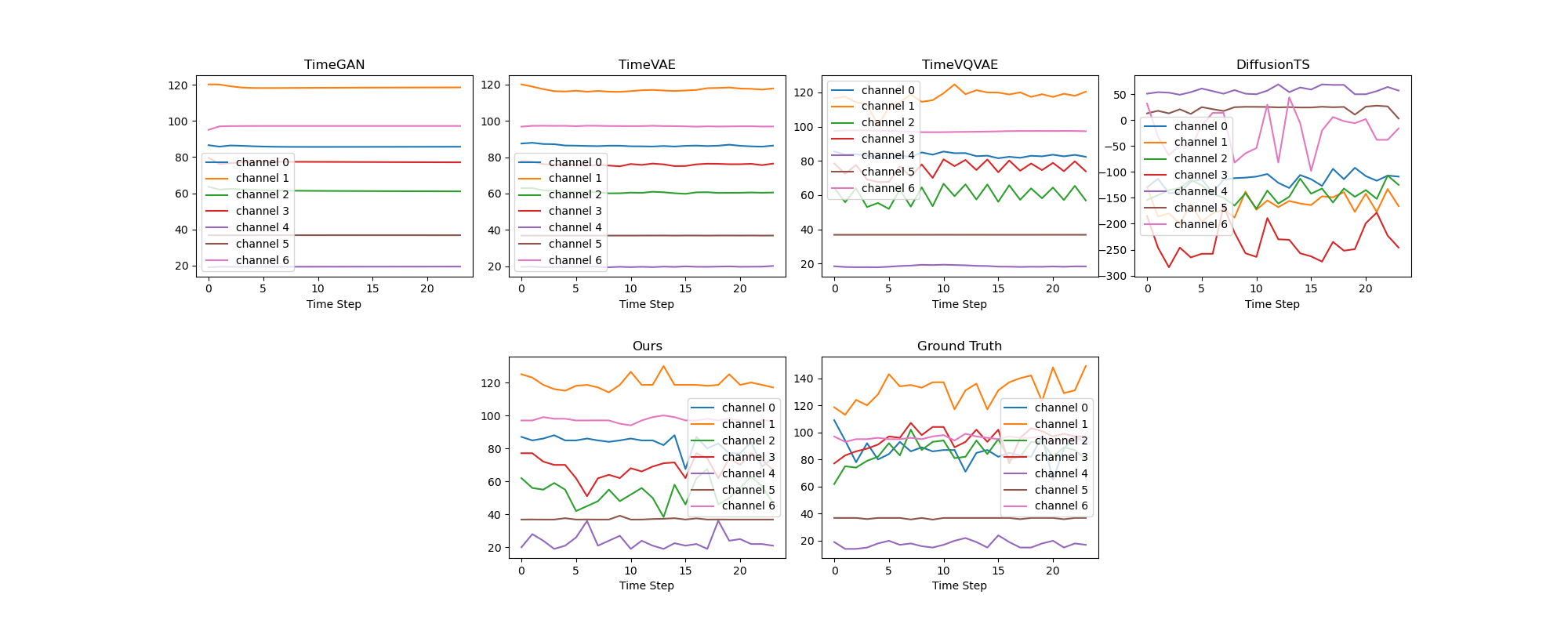}
    }
    \caption{Visualization of negative samples generated by different methods for ICU Stay on MIMIC-III}
    \label{mimic_icustay_neg}
\end{figure*}

\begin{figure*}
    \centering
    \scalebox{0.8}{
    \includegraphics[width=\textwidth]{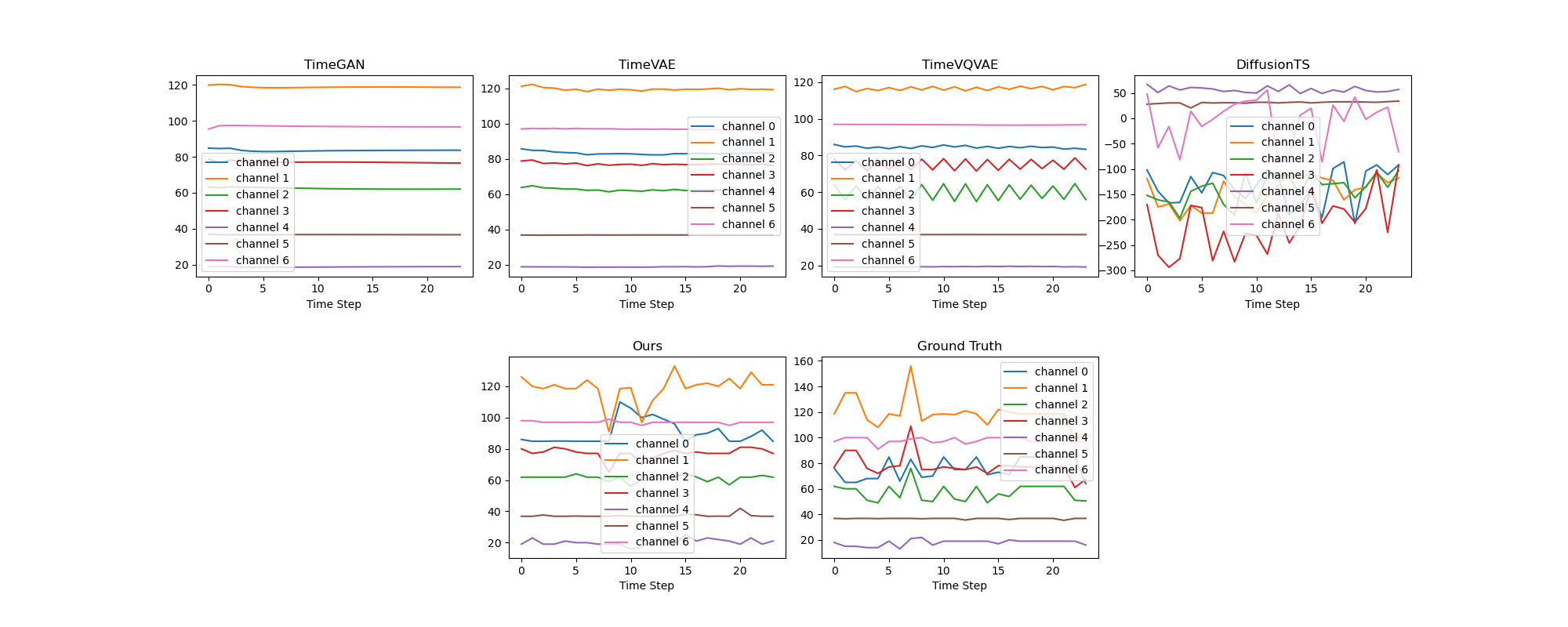}
    }
    \caption{Visualization of positive samples generated by different methods for ICU Stay on MIMIC-III}
    \label{mimic_icustay_pos}
\end{figure*}

\begin{figure*}
    \centering
    \scalebox{0.8}{
    \includegraphics[width=\textwidth]{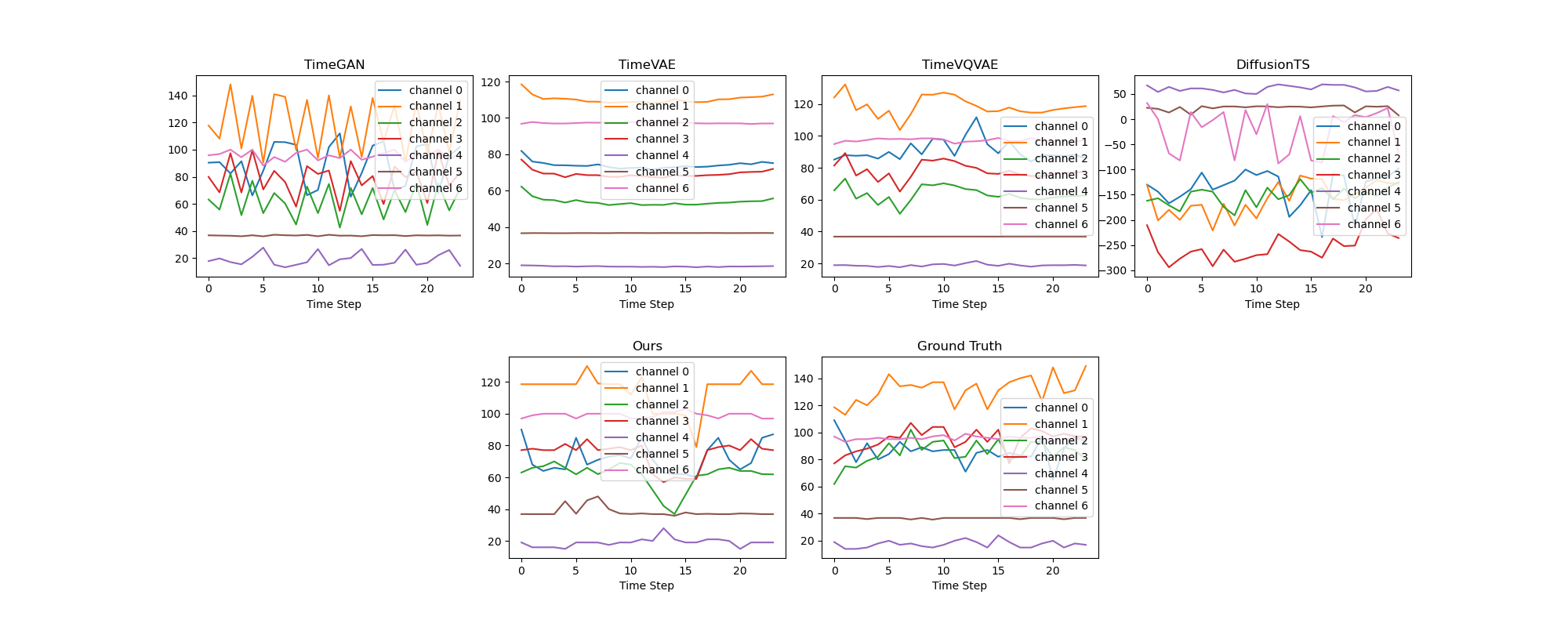}
    }
    \caption{Visualization of negative samples generated by different methods for Mortality  on MIMIC-III}
    \label{mimic_motality_neg}
\end{figure*}

\begin{figure*}
    \centering
    \scalebox{0.8}{
    \includegraphics[width=\textwidth]{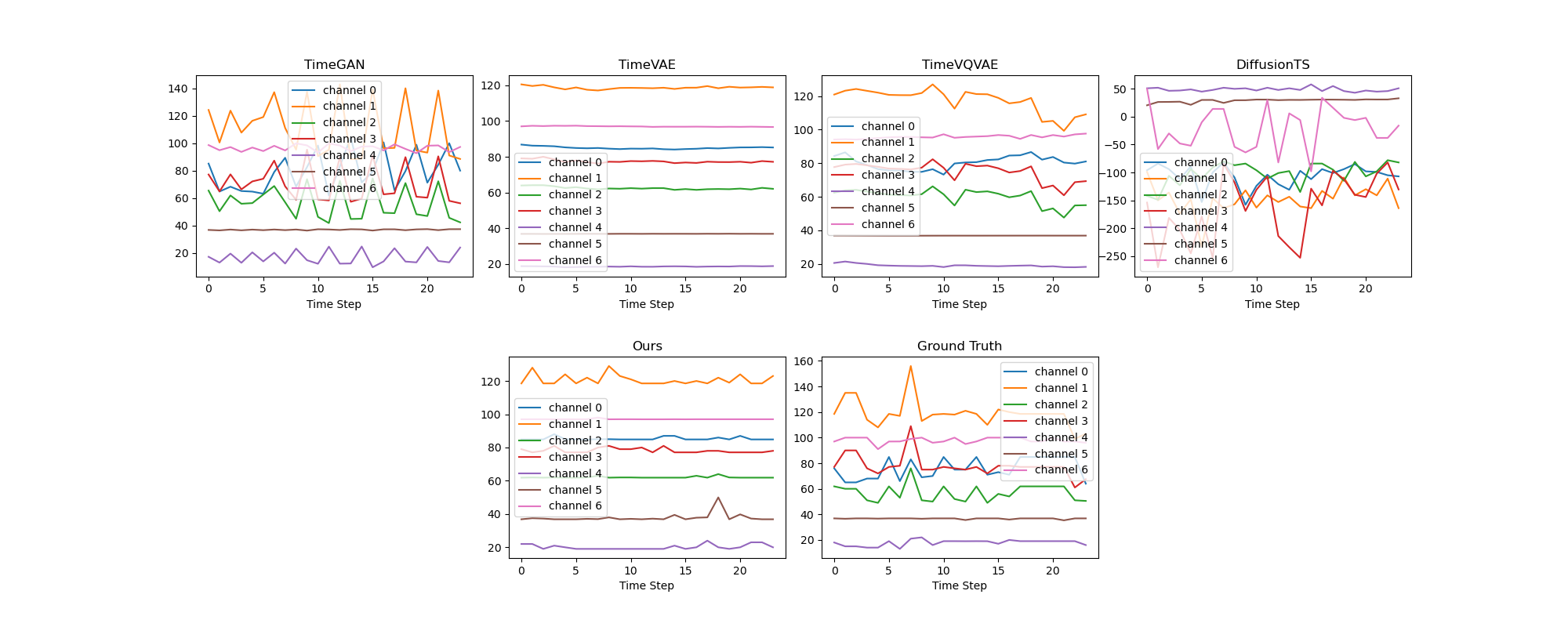}
    }
    \caption{Visualization of positive samples generated by different methods for Mortality on MIMIC-III}
    \label{mimic_motality_pos}
\end{figure*}

\subsection{More TSRTS Results}